\documentclass{article}

\usepackage{arxiv_preprint,times}

\usepackage[utf8]{inputenc} 
\usepackage[T1]{fontenc}    
\usepackage{url}            
\usepackage{booktabs}       
\usepackage{amsfonts}       
\usepackage{nicefrac}       
\usepackage{microtype}      
\usepackage{xcolor}         
\usepackage{caption}
\usepackage{graphicx}
\usepackage{amsmath}

\renewcommand{\cite}{\citep}      
\usepackage{multirow}
\usepackage[most]{tcolorbox}
\usepackage{xcolor}
\usepackage{subcaption}

\definecolor{citecolor}{HTML}{0071bc}
\definecolor{fuchsia}{rgb}{0.6, 0.33, 0.73}
\usepackage[pagebackref=false,breaklinks=true,colorlinks,citecolor=citecolor,bookmarks=false]{hyperref}
\usepackage[capitalize]{cleveref}

\usepackage[table]{xcolor}

\definecolor{aliceblue}{rgb}{0.94,0.97,1.0}

\usepackage{fontawesome5}

\newtcolorbox{boxK}{
    top=2.2pt,
    bottom=2.2pt,
    left=4.5pt,
    right=4.5pt,
    boxrule = 0pt,
    toprule = 0pt, 
}

\newtcolorbox{takeaway}{
  enhanced, breakable,
  colback=gray!5, colframe=gray!5,
  borderline west={2pt}{0pt}{black!70},
  boxrule=0pt, arc=2pt,
  left=10pt, right=8pt, top=6pt, bottom=6pt,
  fontupper=\small
}

\title{$\boldsymbol{\mathcal{D}^2}$-Monitor: $\boldsymbol{\mathcal{D}}$ynamic Safety Monitoring for $\boldsymbol{\mathcal{D}}$iffusion LLMs via Hesitation Signals}

\author{
\textbf{Aoxi Liu}\textsuperscript{1,2,*} \quad
\textbf{Yupeng Chen}\textsuperscript{1,*} \quad
\textbf{James Oldfield}\textsuperscript{1} \\[3pt]
\textbf{Guanzhe Hong}\textsuperscript{1} \quad
\textbf{Junchi Yu}\textsuperscript{1} \quad
\textbf{Baoyuan Wu}\textsuperscript{2} \\[3pt]
\textbf{Philip Torr}\textsuperscript{1} \quad
\textbf{Adel Bibi}\textsuperscript{1,\ensuremath{\dagger}} \\[7pt]
\textsuperscript{1}Torr Vision Group, University of Oxford\\
\textsuperscript{2}The Chinese University of Hong Kong, Shenzhen
}

\begin{document}
\ificlrfinal
\renewcommand{\thefootnote}{\fnsymbol{footnote}}
\footnotetext[1]{Equal contribution.}
\footnotetext[2]{Corresponding author. Email: \texttt{adel.bibi@eng.ox.ac.uk}.}
\renewcommand{\thefootnote}{\arabic{footnote}}
\fi

\maketitle

\begin{abstract}

Despite the emergence of diffusion large language models (D-LLMs) as an alternative to autoregressive large language models (AR-LLMs), safety monitoring for D-LLMs remains largely unexplored. Unlike AR-LLMs, D-LLMs generate text through a \textit{multi-step} denoising process, exposing intermediate hidden representations that may contain safety-relevant information unavailable in standard \textit{single-step} monitoring setups. 
Motivated by the suitability of lightweight probes for always-on monitoring, we analyze which trajectory-level signals best indicate when such probes are likely to struggle. We find that the most informative signal is \textit{safety hesitation}: intermediate hidden states repeatedly falling within a small margin of the probe's decision boundary. 
The number of such hesitation steps in D-LLM's trajectory predicts probe failure effectively, providing a proxy for sample difficulty. 
Building on this analysis, we propose $\boldsymbol{\mathcal{D}^2}$\textbf{-Monitor},
in which a lightweight probe runs on every sample, producing a base
prediction and a hesitation estimate that decides whether to escalate the
sample to a more expressive probe. That probe is trained on trajectories with at least one hesitation step, and
reads only the minimal span of those steps.
Evaluated on 3 datasets (WildGuardMix, ToxicChat, OpenAI-Moderation) across 4 D-LLMs, $\mathcal{D}^2$-Monitor achieves state-of-the-art performance with a compact parameter footprint ($\leq$  0.93M parameters), and exhibits the best trade-off between effectiveness and efficiency relative to 8 baselines.   


\end{abstract}

\section{Introduction}
\label{intro}
Building on causal attention \cite{vaswani2017attention} and the next-token prediction paradigm, autoregressive large language models (AR-LLMs) \cite{achiam2023gpt, grattafiori2024llama, yang2025qwen3} have achieved remarkable performance across diverse tasks, including code generation \cite{chen2021evaluating, li2022competition} and mathematical reasoning \cite{cobbe2021training, wei2022chain}.
Despite their success, this autoregressive paradigm introduces inherent limitations: the sequential decoding constrains generation efficiency and prevents models from revising earlier outputs in light of future context.
Diffusion large language models (D-LLMs) \cite{nie2025large, zhu2025llada, bie2025llada2, labs2025mercury} have recently emerged as a promising alternative. 
Rather than generating tokens sequentially, D-LLMs iteratively refine the entire sequence through a denoising process with bidirectional attention \cite{sahoo2024simple, shi2024simplified}, enabling faster and more flexible generation.
The commercial D-LLM \texttt{Mercury 2} \cite{mercury2} generates 1009 tokens per
second, significantly outperforming AR-LLMs such as \texttt{Claude Haiku 4.5}
(89 tokens/sec) and \texttt{GPT-5-mini} (71 tokens/sec). On the open-source side,
\texttt{LLaDA 2.0} \cite{bie2025llada2} scales D-LLMs to 100B parameters and rivals leading AR-LLMs \cite{yang2025qwen3}.

Despite these advances, safety monitoring for D-LLMs remains underexplored. Effective monitoring is critical: frontier large language models already significantly lower the barrier for malicious actors to execute harmful tasks \cite{anthropic2025disrupting}.
Initial work on D-LLM safety has focused primarily on alignment techniques \cite{jeung2026ad, li2026diffuguard} that improve safety awareness within the model. However, alignment alone is insufficient, as such techniques remain vulnerable to adversarial attacks \cite{nasr2025attacker}. We therefore focus on \textit{external} safety monitors in this paper, which are deployment-time systems that detect harmful user inputs \cite{jiao2026llm,oldfield2026beyond} or problematic model behaviors \cite{goldowsky2025detecting, mckenzie2025detecting}. 

Existing safety monitoring literature has focused on AR-LLMs and falls into two broad categories. \textit{LLMs-as-monitors} \cite{inan2023llama, zeng2025shieldgemma} employ additional LLMs to classify the safety of user prompts or model responses. \textit{Probe-based monitors} operate on internal model representations, which have been shown to encode rich semantic information \cite{alain2016understanding, meng2022locating}. Their lightweight architectures suit always-on, low-cost deployment, and are increasingly adopted in production systems such as Google’s \texttt{Gemini} \cite{kramar2026building}. 


\begin{figure*}[t]
    \centering
    \includegraphics[width=1.0\textwidth]{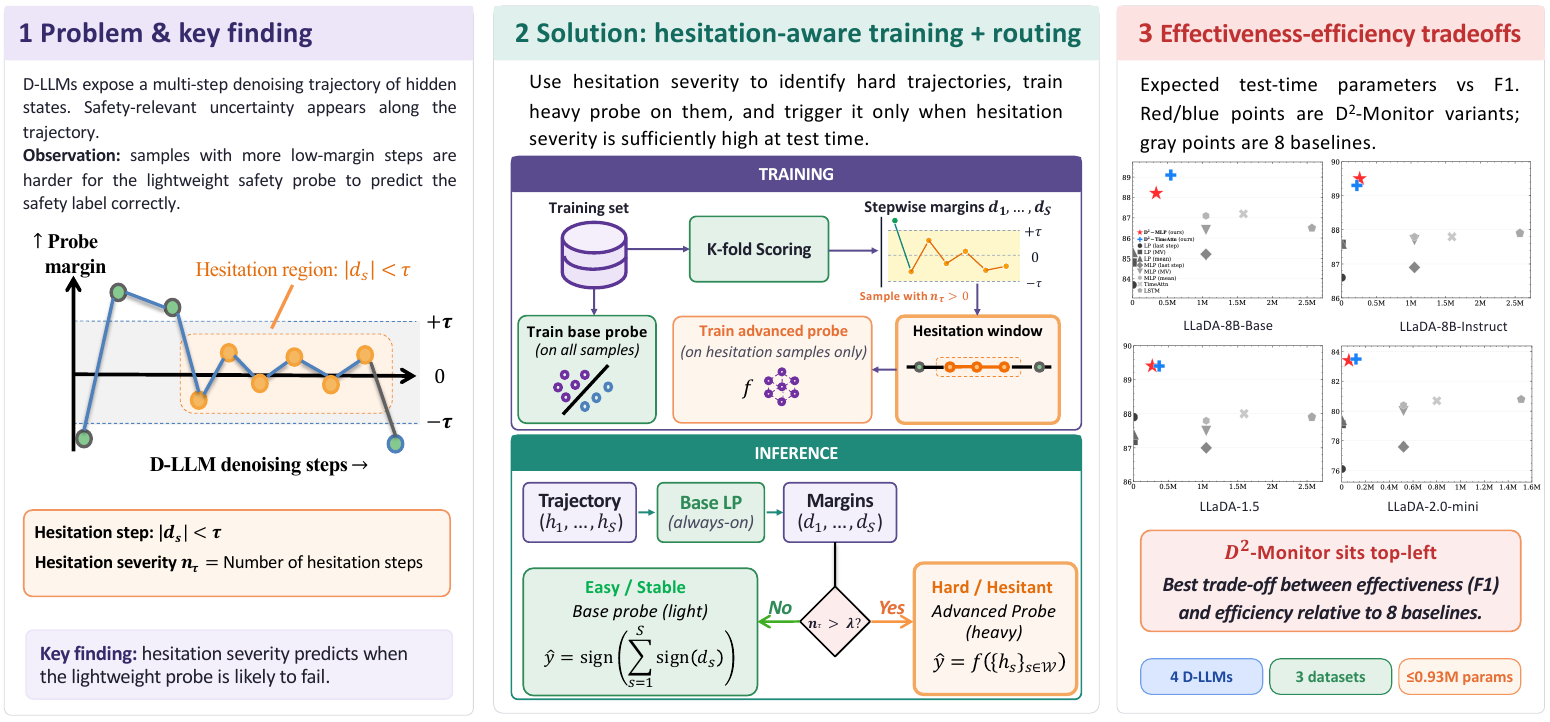}
    \caption{\textbf{Left}: Hesitation severity $n_\tau$ counts the denoising steps whose probe margin falls within $\tau$, and predicts when the lightweight probe fails. \textbf{Middle}: $\mathcal{D}^2$-Monitor uses $n_\tau$ to curate the advanced probe's training data and to route at test time. \textbf{Right}: Expected test-time parameters versus F1 on \textit{WildGuardMix}; $\mathcal{D}^2$-Monitor attains the best F1 with fewer parameters than most baselines.}
    \label{fig:pareto}
\end{figure*}

In this paper, we first argue that D-LLMs' \textit{multi-step} trajectory provides a richer and more useful signal for safety monitoring than \textit{single-step} representations (\cref{sec:trajectory-as-signal}). Inspired by recent findings that intermediate D-LLM outputs can oscillate between correct and incorrect answers during mathematical reasoning \cite{wang2026time, li2026diffusion}, we show that analogous instability occurs in the safety probe space. Specifically, we identify
\textit{hesitation steps}, i.e., intermediate denoising steps whose representations lie close to the probe decision boundary (\cref{sec:trajectory-hesitation}). We further demonstrate that trajectories with more hesitation steps are harder for probes to classify correctly. This establishes hesitation as an effective proxy for sample difficulty, and naturally motivates a bi-level monitor design that routes hard samples to a high-complexity probe while processing easy samples with a lightweight one, dynamically allocating computational resources at test time.

\paragraph{Proposed Work} We introduce $\boldsymbol{\mathcal{D}^2}$\textbf{-Monitor}, a dynamic safety monitor for D-LLMs that harnesses intrinsic safety hesitation in the multi-step denoising trajectory. 
Following prior work on cascaded monitors~(\cref{cascaded design}),
$\mathcal{D}^2$-Monitor adopts a bi-level design, in which we use the hesitation severity $n_\tau$ as the routing signal and the minimal span of hesitation steps to train the more expressive probe.
The base probe serves as an \textit{always-on monitor}, jointly estimating hesitation and performing base-level safety classification. When the hesitation level exceeds a threshold, the router activates the high-complexity advanced probe for second-stage classification.
This dynamic routing mechanism allocates monitoring resources efficiently: easy samples incur only lightweight compute cost, while harder samples (such as adversarially crafted inputs) trigger additional safeguards, achieving a practical balance between effectiveness and efficiency.

We evaluate $\mathcal{D}^2$-Monitor on 3 safety datasets (WildGuardMix, ToxicChat, OpenAI-Moderation) across 4 D-LLMs under both \textit{intra-dataset} and \textit{cross-dataset} settings.
$\mathcal{D}^2$-Monitor achieves state-of-the-art performance with an extremely compact parameter footprint (fewer than 0.93M parameters, or 0.01\%
of an 8B model), and exhibits the best trade-off between efficiency and effectiveness relative to 8 baselines. Additional analysis confirms robustness across generation configurations, remasking strategies, and hyperparameter settings. 
Our main contributions are threefold:
\begin{itemize}
\item We characterize \textit{safety hesitation} in the multi-step hidden states of D-LLMs using probe margins, and show that hesitation severity strongly correlates with linear probe performance.
\item We introduce $\boldsymbol{\mathcal{D}^2}$\textbf{-Monitor}, a dynamic safety monitor for D-LLMs that uses trajectory-level hesitation signals both for curating advanced probe training data and for test-time routing.
\item Tested on 3 safety datasets across 4 D-LLMs, $\mathcal{D}^2$-Monitor achieves state-of-the-art performance under both \textit{intra-dataset} and \textit{cross-dataset} settings, with the best trade-off between effectiveness and efficiency against 8 baselines.
\end{itemize}
    

\section{Related Work}
\label{related}

\paragraph{Diffusion Large Language Models} Diffusion large language models
(D-LLMs) \cite{nie2025large, zhu2025llada, bie2025llada2}, built upon masked
diffusion models \cite{austin2021structured, hoogeboom2021argmax,
shi2024simplified, nie2025scaling}, extend diffusion-based generative modeling
from continuous domains such as images \cite{yang2023diffusion} to discrete
text. Recent work has identified intrinsic safety-relevant properties of
diffusion-based generation relative to autoregressive generation
\cite{he2026safer}. Early efforts to safeguard D-LLMs have mainly explored
finetuning-based defenses \cite{jeung2026ad} and decoding intervention
\cite{li2026diffuguard}. However, finetuning approaches incur substantial computational overhead and may affect model utility, while decoding intervention requires regeneration, which affects efficiency. We instead adopt a probe-based monitoring approach: a lightweight auxiliary module that can be deployed alongside any D-LLM without modifying the underlying model, offering a practical and non-intrusive defense mechanism.

\paragraph{LLM Monitors} Despite extensive safety training, LLMs remain vulnerable to adversarial attacks \cite{liu2024autodan, zeng2024johnny, chen2026alignment}, making external safety guardrails necessary
\cite{kramar2026building}. These guardrails fall into two broad categories.
\textbf{(1) LLMs-as-monitors.} One approach deploys an additional LLM trained as a safety classifier to filter inputs and outputs
\cite{openai2022moderation, han2024wildguard, padhi2024granite, liu2025guardreasoner}.
Representative models such as Llama-Guard \cite{inan2023llama} are fine-tuned on safety tasks to improve detection of adversarially crafted prompts. While capable, LLM-based monitors add substantial compute, which is prohibitive in resource-constrained settings such as edge deployment.
\textbf{(2) Probe-based monitors.} A more efficient alternative trains lightweight probes on the model's internal representations, which encode rich
semantic information \cite{park2024linear, jiao2026llm}. 
Linear probes \cite{alain2016understanding} are the canonical example, effective on hallucination detection \cite{han2025simple},
toxicity detection \cite{hu2024toxicity}, and deception detection
\cite{goldowsky2025detecting, macdiarmidsimple}.
More expressive probes range from
bilinear and MLP variants \cite{hewitt2019designing, pimentel2020pareto} to
architectures that read the entire sequence of states produced during generation,
such as temporal attention \cite{kramar2026building} and recurrent encoders
\cite{damirchi2026truth}. For AR-LLMs, each generated token contributes one state, whereas a D-LLM re-encodes the whole sequence at every denoising step,
producing a trajectory of intermediate states~(\cref{sec:problem}).

\paragraph{Cascaded Design} 
\label{cascaded design}
Routing easy inputs to a cheap model and
escalating hard ones is a long-standing idea, from classical detection cascades
\cite{viola2004robust} to early-exit networks
\cite{teerapittayanon2016branchynet}. Recent work applies this to LLM
monitoring. 
\citet{mckenzie2025detecting} and \citet{cunningham2026constitutional} use an
activation probe as a first stage and defer uncertain inputs to an external
LLM-as-monitor, reporting substantial savings over running that monitor on every input. 
The cost is that the second
stage remains an LLM: it must be finetuned or prompted in advance, and it adds a
model call at inference time. 
Our work differs in three key respects: (1) we introduce a D-LLM specific multi-step routing signal, (2) we use a probe as second-stage classifier instead of an additional LLM, and (3) we score training samples to curate hesitation trajectories for training the second-stage probe. We do not claim the bi-level design itself
as a novelty.
\section{Exploring Safety Monitoring in D-LLMs}
\label{sec:trajectory-analysis}

\subsection{Preliminary}

\paragraph{Diffusion Large Language Models}
Diffusion large language models define a discrete diffusion process over token sequences. 
Let $\mathbf{x}^{(1)} \in \mathcal{V}^L$ denote a clean text sequence, where $\mathcal{V}$ is the vocabulary and $L$ is the sequence length. 
The forward noising process gradually corrupts $\mathbf{x}^{(1)}$ into noisy states $\mathbf{x}^{(2)}, \ldots, \mathbf{x}^{(S)}$, where $\mathbf{x}^{(S)}$ is a fully masked sequence. 
This process is specified by a fixed corruption distribution $q(\mathbf{x}^{(2:S)} \mid \mathbf{x}^{(1)}) = \prod_{s=2}^{S} q(\mathbf{x}^{(s)} \mid \mathbf{x}^{(s-1)})$, where $q(\mathbf{x}^{(s)} \mid \mathbf{x}^{(s-1)})$ masks tokens according to a predefined noise schedule.

The reverse process is parameterized as $p_\theta(\mathbf{x}^{(1:S)}) = p(\mathbf{x}^{(S)}) \prod_{s=2}^{S} p_\theta(\mathbf{x}^{(s-1)} \mid \mathbf{x}^{(s)})$.
At each reverse step, the model samples predictions for the whole sequence from $p_\theta(\mathbf{x}^{(1)} \mid \mathbf{x}^{(s)})$, but only replaces the currently masked positions with the predicted tokens. 
The next state $\mathbf{x}^{(s-1)}$ is then constructed by remasking a fraction $\rho_s$ of the newly predicted positions according to a remasking strategy,
such as random or low-confidence remasking~\cite{nie2025large}.
In practice, given a prompt, the reverse denoising process starts from a partially masked state $\tilde{\mathbf{x}}^{(S)}$, obtained by placing the prompt as a fixed unmasked prefix in $\mathbf{x}^{(S)}$ while keeping the remaining positions masked.

\paragraph{Problem Setup}
\label{sec:problem}

Given a dataset of \(I\) prompts with safety labels \(y^{(i)} \in \{0,1\}\) indicating whether the i-th prompt is safe ($0$) or unsafe ($1$),
the D-LLM produces a hidden representation \(\mathbf{H}^{(i)}_{\mathrm{raw}} \in \mathbb{R}^{D \times L \times S}\) for the i-th prompt at a particular layer, where \(D\), \(L\), and \(S\) denote the hidden dimension, sequence length, and number of denoising steps respectively. 
Since D-LLMs adopt bidirectional attention, safety-relevant information is distributed across tokens. 
We therefore aggregate over the sequence dimension via mean pooling, yielding a step-wise representation matrix
\(\mathbf{H}^{(i)}=[\mathbf{h}^{(i)}_1,\ldots,\mathbf{h}^{(i)}_S]\in \mathbb{R}^{D \times S}\), 
where \(\mathbf{h}^{(i)}_s\in\mathbb{R}^{D}\) denotes the aggregated hidden state at step \(s\). 
The dataset of \(I\) representation matrices and their labels is denoted by \(\mathcal{D} = \{\mathbf{H}^{(i)}, y^{(i)}\}^{I}_{i=1}\).
A safety probe \(f\) is learned by minimizing the empirical cross-entropy loss:
\begin{equation}
    \min_{f} 
    \frac{1}{I}\sum_{i=1}^{I}
    \mathcal{L}\!\left(y^{(i)}, f(\mathbf{H}^{(i)})\right).
    \label{eq:probe-objective}
\end{equation}
We instantiate $f$ as a linear probe, as its lightweight design makes it suitable for always-on monitoring while maintaining strong interpretability \citep{hewitt2019designing}.

\subsection{Multi-step as Useful Signal: Beyond Single-Step Safety Probing}
\label{sec:trajectory-as-signal}


The first question to consider in optimizing a probe to monitor D-LLMs is the choice of $\mathbf{H}$ due to the richer multi-step hidden representations compared to AR-LLMs. 
Specifically, should a probe rely on a single-step representation \(\mathbf{h}_s\), or does the full trajectory \(\mathbf{H}\) carry additional safety-relevant signal?

To answer this, we compare two monitoring settings that differ in how \(\mathbf{H}\) is used.
\textbf{(1) Single-step probing:}
The probe operates on a single denoising step. We use the final-step representation \(\mathbf{h}_1\), since it is the most refined hidden state before generation terminates, and train and test the probe on \(\mathbf{h}_1\).
\textbf{(2) Multi-step probing:}
The probe operates on the full denoising trajectory \(\mathbf{H}\). To keep training cost comparable to the single-step setting, we train on the temporal-mean representation
\(\bar{\mathbf{h}}=\frac{1}{S}\sum_{s=1}^{S}\mathbf{h}_s\)
rather than treating each denoising step as a separate training example. At test time, we consider two trajectory-level readouts: 
\textit{(a) Mean}, which evaluates the probe directly on \(\bar{\mathbf{h}}\), i.e., \(f(\bar{\mathbf{h}})\); and \textit{(b) Majority Vote (MV)}, which applies the same probe to each individual step and aggregates via majority voting:
\begin{equation}
    \hat{y}^{(i)}
    =
    \mathrm{Majority}
    \left(
    f(\mathbf{h}^{(i)}_1),\ldots,f(\mathbf{h}^{(i)}_S)
    \right).
    \label{eq:majority-voting}
\end{equation}
Both multi-step readouts use the same probe setting and number of training samples as the single-step setting, enabling a controlled comparison of trajectory utilization.


We denote the three settings by \textbf{LP (Last Step)}, \textbf{LP (Mean)}, and
\textbf{LP (MV)} (Appendix~\ref{app:probe_architectures}).
As shown in \cref{tab:main_results,tab:main_results_toxicchat}, both multi-step readouts achieve higher Acc and F1 than the single-step baseline on most models, indicating that intermediate denoising steps carry safety-relevant information that the final step alone misses. 
We therefore use the full trajectory \(\mathbf{H}\) for all subsequent analysis.

\subsection{Hesitation Steps as Difficulty Signal: Separating Easy and Hard Samples}
\label{sec:trajectory-hesitation}






Despite linear probes' low cost and interpretable form, they have limited expressivity and may fail to capture non-linear structure in representations~\citep{10.1162/coli_a_00422, white2021non}, leading to misclassification on ``harder'' samples.
We therefore seek signals that reflect when a linear probe is likely to struggle.
Inspired by recent findings that intermediate D-LLM responses can fluctuate between correct and incorrect answers during mathematical reasoning~\cite{wang2026time, li2026diffusion}, we hypothesize that analogous instability may occur in the safety context: the model may exhibit uncertainty in its safety decisions across the denoising trajectory. Accordingly, trajectories may be characterized as \emph{stable}, where the model remains consistent across steps, and \emph{hesitant}, where high uncertainty arises at intermediate steps.

\begin{figure}[t]
    \centering

    \begin{subfigure}[t]{0.24\linewidth}
        \centering
        \includegraphics[width=\linewidth]{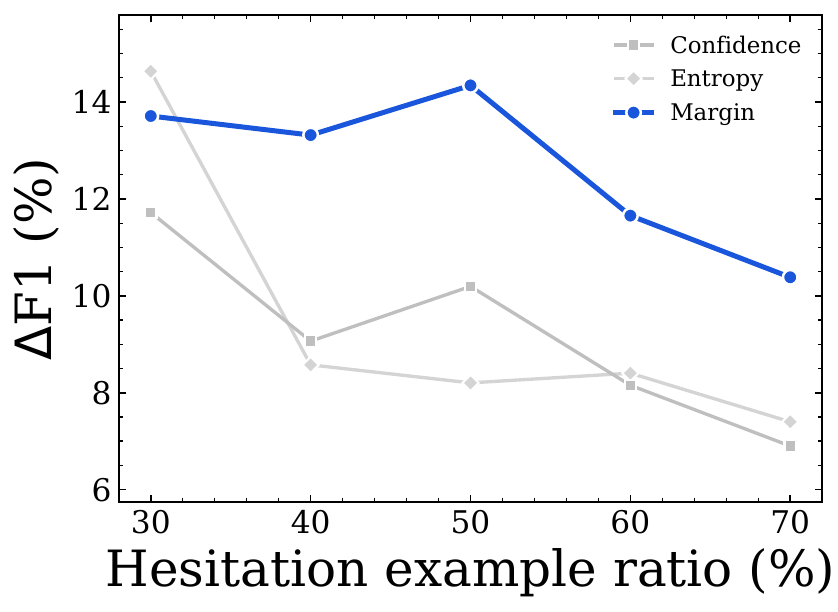}
        \caption{$\Delta$F1 - LP (MV)}
        \label{fig:combined_probes:a}
    \end{subfigure}
    \hfill
    \begin{subfigure}[t]{0.24\linewidth}
        \centering
        \includegraphics[width=\linewidth]{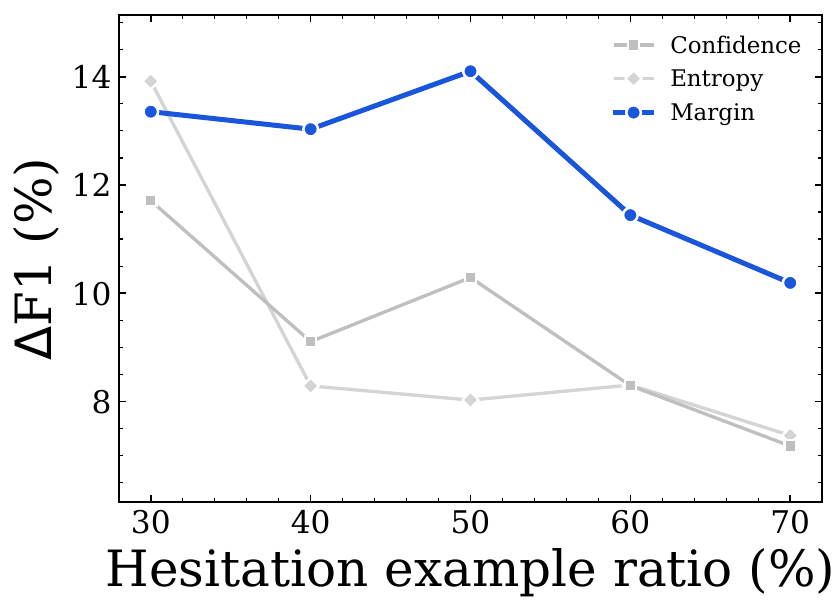}
        \caption{$\Delta$F1 - LP (Mean)}
        \label{fig:combined_probes:b}
    \end{subfigure}
    \hfill
    \begin{subfigure}[t]{0.24\linewidth}
        \centering
        \includegraphics[width=\linewidth]{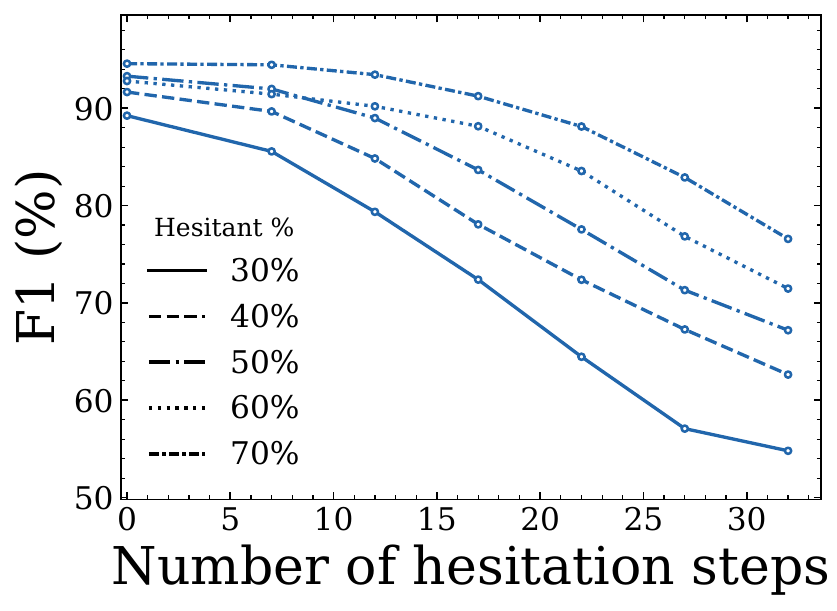}
        \caption{F1 - LP (MV)}
        \label{fig:combined_probes:c}
    \end{subfigure}
    \hfill
    \begin{subfigure}[t]{0.24\linewidth}
        \centering
        \includegraphics[width=\linewidth]{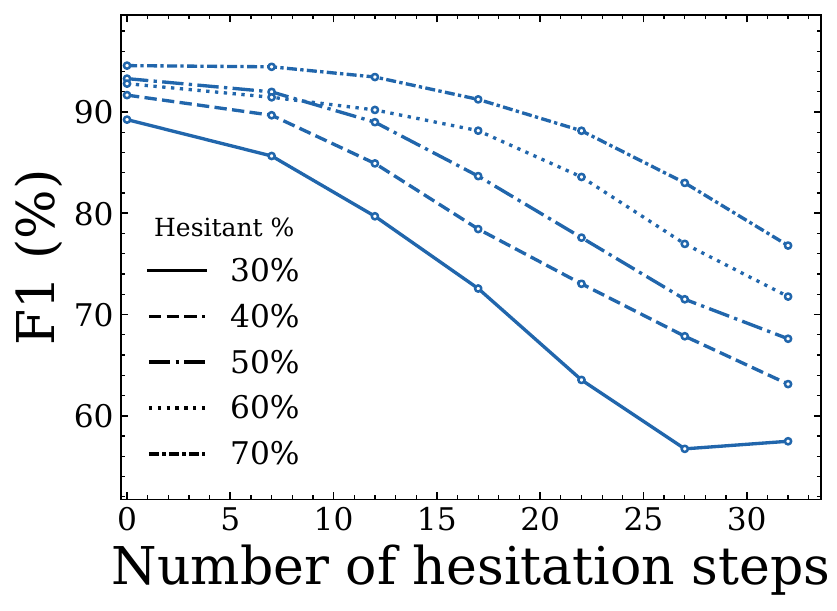}
        \caption{F1 - LP (Mean)}
        \label{fig:combined_probes:d}
    \end{subfigure}

    \caption{
    \textbf{(a)(b):} F1 differences across probing methods under varying ratios of hesitation examples.
    \textbf{(c)(d):} F1 score as a function of the number of hesitation steps under different threshold values $\tau$.
    }
    \label{fig:combined_probes}
\end{figure}

\paragraph{Hesitation Characterization}
To verify this hypothesis, we explore two types of signals that may inform on such hesitation.
\textbf{(1) Probe-extrinsic signals}
quantify uncertainty from the model's predicted token distribution, independently of the probe.
Let \(\mathcal{R}\) denote the set of sequence positions and \(p^{(r,v)}_s\) the predicted probability of token \(v\) at position \(r\) and denoising step \(s\).
We define the step-wise entropy score $E_s$ and confidence score $C_s$ as 
\begin{equation}
    E_s = -|\mathcal{R}|^{-1} \sum_{r\in\mathcal{R}} \sum_v p^{(r,v)}_s \log p^{(r,v)}_s, \quad C_s = |\mathcal{R}|^{-1} \sum_{r\in\mathcal{R}} \max_v p^{(r,v)}_s.
\end{equation}
A step is flagged as hesitant if \(E_s \geq \tau_E\) or \(C_s \leq \tau_C\) for thresholds \(\tau_E\) and \(\tau_C\). 
A trajectory is considered hesitant if it contains at least one hesitation step.
\textbf{(2) Probe-intrinsic signals}
measure uncertainty with respect to the probe's decision boundary. 
Applying the linear probe from \cref{sec:trajectory-as-signal} to each \(\mathbf{h}_s\) yields a step-wise logit.
Let \(d_s\) denote the signed margin to the decision boundary. A step is flagged as hesitant if \(|d_s| < \tau\) for a margin threshold \(\tau\). 
A trajectory is considered hesitant if at least one of its steps is hesitant.
We then compare probe performance on the stable and hesitant subsets characterized by these two kinds of signals across a range of thresholds.
For a fair comparison, the thresholds are chosen to produce comparable split ratios between the two subsets. As shown in \cref{fig:combined_probes:a,fig:combined_probes:b},
probe performance differs substantially: 
\textit{hesitation trajectories yield markedly lower F1 scores than stable ones, confirming that trajectory hesitation is predictive of classification difficulty}. Among the signals evaluated, \textit{the probe margin produces the largest performance gap, indicating that probe-intrinsic signals most effectively identify hard trajectories.}
In Appendix~\ref{app:hesitation_analysis}, we further conduct a dynamical analysis to understand the underlying mechanism.

\paragraph{Hesitation Severity}
\label{hesitaion severity}
However, this $\tau$-induced criterion only captures whether a trajectory exhibits hesitation, not its extent.
We therefore define hesitation severity
\(n_\tau = \sum_{s=1}^{S} \mathbb{I}[|d_s| < \tau]\),
under which the original criterion is simply $\mathbb{I}[n_\tau \geq 1]$.
Comparing probe F1 under the two partitions
(\cref{fig:combined_probes:a,fig:combined_probes:b,fig:combined_probes:c,fig:combined_probes:d})
yields two observations.
\textbf{(1) $\boldsymbol{n_\tau}$ gives finer levels and a wider F1 gap.}
The binary partition into stable ($n_\tau=0$) and hesitant ($n_\tau\geq1$)
trajectories yields an F1 gap of 10--14 points under the margin signal
(\cref{fig:combined_probes:a,fig:combined_probes:b}). When we stratify by $n_\tau$ instead, probe F1 decreases monotonically with $n_\tau$, and the gap between
the extreme buckets reaches $\sim$30 points at the 30\% hesitant example ratio
(\cref{fig:combined_probes:c,fig:combined_probes:d}).
\textbf{(2) $\boldsymbol{\tau}$ over-flags easy trajectories.}
Probe F1 on small-$n_\tau$ trajectories remains close to that on the stable
subset, yet the
binary criterion lumps them together with those genuinely hard for the probe~(\cref{fig:combined_probes:c,fig:combined_probes:d}).
Defining $n_{\text{entropy}}$ and $n_{\text{confidence}}$ analogously from
probe-extrinsic signals, we find $n_\tau$ the most predictive of difficulty
(Appendix \ref{app:step_count_signals}).

\section{$\boldsymbol{\mathcal{D}^2}$\textbf{-Monitor}}
\label{method}
\subsection{overview}

Inspired by prior work on hierarchical monitoring in AR-LLMs
\citep{mckenzie2025detecting, cunningham2026constitutional, oldfield2026beyond}
and by our finding that $n_\tau$ predicts base-probe failure
(\cref{sec:trajectory-hesitation}), we propose $\mathcal{D}^2$-Monitor, a monitoring framework in which a
low-complexity \emph{base probe} scores every sample and a \emph{router}
escalates the hard ones to a high-complexity \emph{advanced probe}
(\cref{fig:pareto}). We make two
main design choices. First, whereas existing cascaded monitors escalate on a
single uncertainty score from the first-stage probe
\citep{mckenzie2025detecting, cunningham2026constitutional}, we escalate on
$n_\tau$, which accumulates evidence across denoising steps. Since $n_\tau$ is
read off the base probe's per-step margins, the rule reduces to a single
threshold test $n_\tau > \lambda$, where $\lambda$ exposes a compute budget:
larger $\lambda$ escalates fewer samples, and performance remains strong even
at low escalation rates~(Appendix~\ref{app:sensitivity}). Second, hesitation steps
tend to occur in a contiguous run rather than scattered across the trajectory
(Appendix~\ref{Margin Persistence}), so we feed the advanced probe only the minimal
span containing them, rather than all $S$ steps; we ablate this choice
in Appendix~\ref{app:window_ablation}.

\subsection{Implementation}

Given its lightweight architecture, strong performance on low-hesitation samples~(\cref{fig:combined_probes:c,fig:combined_probes:d}), and effectiveness at identifying estimation difficulty (Section~\ref{sec:trajectory-hesitation}), we adopt the linear probe as the low-complexity base probe.
Our framework is flexible with respect to the choice of the high-complexity probe. In this work, we consider two variants with comparable parameter counts: (1) an MLP probe and (2) a temporal attention probe (TimeAttn) that aggregates hidden states within the hesitation window. Additional architectural details are provided in Appendix~\ref{app:probe_architectures}.

\paragraph{Stage 1: Out-of-Fold Scoring and Hesitation Trajectories Collection}
We first identify which training trajectories hesitate. To obtain unbiased estimates, we score the training set out-of-fold (OOF): it is split
into $k$ folds, and each fold is scored by a linear probe trained on the
remaining $k-1$, yielding margins $\{d_s^{(i)}\}_{s=1}^{S}$ for every training example
$i$. Counting the steps with $|d_s^{(i)}|<\tau$ gives $n_\tau^{(i)}$, and we select
trajectories with $n_\tau^{(i)}>0$ for training the advanced probe.

\paragraph{Stage 2: Base and Advanced Probes Training}
We train the base linear probe on the full training set, which is used at test time to compute step-wise margins and derive the hesitation severity \(n_\tau\).
For the high-complexity probe, we define the hesitation window $\mathcal{W}^{(i)}$ as
the minimal contiguous span containing the hesitation steps of trajectory $i$.
It is empty when the trajectory does not hesitate. We train the advanced probe $f$ on the
non-empty windows:


\begin{equation}
\min_{f} \sum_{i:\, n_\tau^{(i)} > 0} \mathcal{L}\!\left(y^{(i)},\; f(\{\mathbf{h}_s^{(i)}\}_{s \in \mathcal{W}^{(i)}})\right),
\end{equation}


\paragraph{Stage 3: Cascade Detection}
At test time, the base probe computes the signed margin $d_s$ at every
denoising step, and $n_\tau$ counts the steps with $|d_s|<\tau$. If
$n_\tau\leq\lambda$, the base probe classifies the example by majority voting:
$\hat{y}=\mathrm{sign}\big(\sum_{s=1}^{S}\mathrm{sign}(d_s)\big)$. Otherwise,
the hesitation window $\mathcal{W}$ is extracted and passed to the advanced
probe, which produces the final prediction. Details on the hyperparameters \(\tau\) and \(\lambda\) are provided in Appendix \ref{sec:appendix_hparam}.
\section{Experiment}
\label{exp}
\subsection{Experiment Setup}

\paragraph{Datasets}
We evaluate on three safety datasets. \textbf{WildGuardMix}~\cite{han2024wildguard} consists of 86.8k training prompts and 1.7k test prompts, each labeled as harmful or unharmful. The dataset includes many adversarially designed inputs, posing a challenging benchmark for safety evaluation. \textbf{ToxicChat}~\cite{lin2023toxicchat} contains 5.08k training and 5.08k test prompts collected from real user-AI interactions, each annotated with a binary toxicity label (1 for toxic, 0 otherwise). \textbf{OpenAI-Moderation}~\cite{openai2022moderation} consists of 1.68k prompts annotated across eight moderation categories such as hate speech, violence, and self-harm; a prompt is labeled as unsafe if any category is flagged, and safe otherwise. For \textit{intra-dataset} evaluation, we train and test on WildGuardMix and ToxicChat separately. For \textit{cross-dataset} evaluation, we train on WildGuardMix and test on both ToxicChat and OpenAI-Moderation.

\paragraph{Models}
We use four open-source D-LLMs for our experiments: \texttt{LLaDA-8B-Base} \cite{nie2025large}, \texttt{LLaDA-8B-Instruct}~\cite{nie2025large}, \texttt{LLaDA-1.5-8B}~\cite{zhu2025llada}, and \texttt{LLaDA-2.0-mini-16B}~\cite{bie2025llada2}.

\begin{table*}[t]
\centering
\caption{\textbf{Intra-dataset performance on the test set.} Monitors are trained and tested on \textit{WildGuardMix}. Mean$\pm$std over seeds 0, 1, 2. Best results are in \textbf{bold}, and second-best are \underline{underlined}. }
\label{tab:main_results}
\resizebox{\textwidth}{!}{%
\begin{tabular}{l c cc cc cc c cc}
\toprule
& & \multicolumn{2}{c}{\textbf{LLaDA-8B-Base}} & \multicolumn{2}{c}{\textbf{LLaDA-8B-Instruct}} & \multicolumn{2}{c}{\textbf{LLaDA-1.5}} & & \multicolumn{2}{c}{\textbf{LLaDA-2.0-mini}} \\
\cmidrule(lr){3-4} \cmidrule(lr){5-6} \cmidrule(lr){7-8} \cmidrule(lr){10-11}
\textbf{Method} & \textbf{E[P]} & Acc & F1 & Acc & F1 & Acc & F1 & \textbf{E[P]} & Acc & F1 \\
\midrule
\textit{Single-step methods} \\[2pt]
\quad LP (Last Step) & 4$\times$10$^{-3}$M & 84.2$\pm$0.2 & 83.7$\pm$0.2 & 87.0$\pm$0.3 & 86.6$\pm$0.3 & 88.2$\pm$0.1 & 87.9$\pm$0.1 & 2$\times$10$^{-3}$M & 77.4$\pm$0.2 & 76.1$\pm$0.2 \\
\quad MLP (Last Step) & 1.05M & 85.6$\pm$0.3 & 85.2$\pm$0.4 & 87.2$\pm$0.5 & 86.9$\pm$0.5 & 87.3$\pm$0.0 & 87.0$\pm$0.0 & 0.52M & 78.5$\pm$0.1 & 77.6$\pm$0.1 \\
\midrule
\textit{Full-trajectory methods} \\[2pt]
\quad LP (MV) & 4$\times$10$^{-3}$M & 85.2$\pm$0.7 & 84.8$\pm$0.7 & 87.8$\pm$0.5 & 87.6$\pm$0.5 & 87.4$\pm$0.1 & 87.2$\pm$0.1 & 2$\times$10$^{-3}$M & 80.1$\pm$0.3 & 79.1$\pm$0.2 \\
\quad LP (Mean) & 4$\times$10$^{-3}$M & 85.7$\pm$1.0 & 85.3$\pm$1.0 & 87.9$\pm$0.4 & 87.6$\pm$0.4 & 87.7$\pm$0.1 & 87.4$\pm$0.1 & 2$\times$10$^{-3}$M & 80.3$\pm$0.2 & 79.4$\pm$0.2 \\
\quad MLP (MV) & 1.05M & 86.8$\pm$0.4 & 86.4$\pm$0.4 & 88.0$\pm$0.4 & 87.7$\pm$0.4 & 87.8$\pm$0.1 & 87.5$\pm$0.0 & 0.52M & 80.9$\pm$0.1 & 80.0$\pm$0.0 \\
\quad MLP (Mean) & 1.05M & 87.5$\pm$0.3 & 87.1$\pm$0.4 & 88.1$\pm$0.4 & 87.8$\pm$0.4 & 88.1$\pm$0.0 & 87.8$\pm$0.0 & 0.52M & 81.2$\pm$0.1 & 80.4$\pm$0.0 \\
\quad TimeAttn \cite{kramar2026building} & 1.59M & 87.6$\pm$0.5 & 87.2$\pm$0.6 & 88.1$\pm$0.2 & 87.8$\pm$0.2 & 88.2$\pm$0.4 & 88.0$\pm$0.5 & 0.80M & 81.7$\pm$0.4 & 80.7$\pm$0.2 \\
\quad LSTM \cite{damirchi2026truth} & 2.57M & 87.0$\pm$0.4 & 86.5$\pm$0.4 & 88.2$\pm$0.4 & 87.9$\pm$0.4 & 88.1$\pm$0.4 & 87.9$\pm$0.4 & 1.51M & 81.7$\pm$0.6 & 80.8$\pm$0.5 \\
\midrule
\rowcolor{aliceblue} \quad  $\boldsymbol{\mathcal{D}^2}$\textbf{-MLP (Ours)} & $\leq$0.34M & \underline{88.5$\pm$0.2} & \underline{88.2$\pm$0.3} & \textbf{89.8$\pm$0.1} & \textbf{89.5$\pm$0.1} & \underline{89.7$\pm$0.1} & \underline{89.4$\pm$0.1} & 0.06M & \underline{84.0$\pm$0.1} & \underline{83.4$\pm$0.1} \\
\rowcolor{aliceblue} \quad  $\boldsymbol{\mathcal{D}^2}$\textbf{-TimeAttn (Ours)} & $\leq$0.55M & \textbf{89.4$\pm$0.3} & \textbf{89.1$\pm$0.4} & \underline{89.6$\pm$0.1} & \underline{89.3$\pm$0.1} & \textbf{89.7$\pm$0.2} & \textbf{89.4$\pm$0.2} & 0.12M & \textbf{84.2$\pm$0.3} & \textbf{83.5$\pm$0.3} \\
\bottomrule
\end{tabular}%
}
\end{table*}

\begin{table*}[t]
\centering
\caption{\textbf{Intra-dataset performance on the test set.} Monitors are trained and tested on \textit{ToxicChat}. Mean$\pm$std over seeds 0, 1, 2. Best results are in \textbf{bold}, and second-best are \underline{underlined}.}
\label{tab:main_results_toxicchat}
\resizebox{\textwidth}{!}{%
\begin{tabular}{l c cc cc cc c cc}
\toprule
& & \multicolumn{2}{c}{\textbf{LLaDA-8B-Base}} & \multicolumn{2}{c}{\textbf{LLaDA-8B-Instruct}} & \multicolumn{2}{c}{\textbf{LLaDA-1.5}} & & \multicolumn{2}{c}{\textbf{LLaDA-2.0-mini}} \\
\cmidrule(lr){3-4} \cmidrule(lr){5-6} \cmidrule(lr){7-8} \cmidrule(lr){10-11}
\textbf{Method} & \textbf{E[P]} & Acc & F1 & Acc & F1 & Acc & F1 & \textbf{E[P]} & Acc & F1 \\
\midrule
\textit{Single-step methods} \\[2pt]
\quad LP (Last Step) & 4$\times$10$^{-3}$M & 89.7$\pm$0.8 & 74.6$\pm$1.3 & 90.3$\pm$0.2 & 76.0$\pm$0.4 & 89.5$\pm$1.2 & 75.0$\pm$1.5 & 2$\times$10$^{-3}$M & 93.6$\pm$0.3 & 79.8$\pm$0.5 \\
\quad MLP (Last Step) & 1.05M & 95.6$\pm$0.1 & 79.9$\pm$0.4 & 96.6$\pm$0.0 & 86.9$\pm$0.0 & 96.3$\pm$0.2 & 85.3$\pm$0.8 & 0.52M & 96.1$\pm$0.1 & 82.1$\pm$0.0 \\
\midrule
\textit{Full-trajectory methods} \\[2pt]
\quad LP (MV) & 4$\times$10$^{-3}$M & 89.7$\pm$1.5 & 75.1$\pm$1.8 & 90.6$\pm$1.3 & 76.6$\pm$1.6 & 90.0$\pm$1.4 & 75.9$\pm$1.9 & 2$\times$10$^{-3}$M & 93.4$\pm$0.3 & 80.0$\pm$0.4 \\
\quad LP (Mean) & 4$\times$10$^{-3}$M & 90.5$\pm$1.3 & 76.3$\pm$1.7 & 90.9$\pm$1.4 & 77.2$\pm$1.8 & 90.1$\pm$1.4 & 76.1$\pm$2.0 & 2$\times$10$^{-3}$M & 93.6$\pm$0.2 & 80.8$\pm$0.3 \\
\quad MLP (MV) & 1.05M & 96.1$\pm$0.1 & 83.8$\pm$0.2 & 96.7$\pm$0.1 & 87.1$\pm$0.5 & 96.8$\pm$0.1 & 87.7$\pm$0.0 & 0.52M & 95.9$\pm$0.4 & 81.5$\pm$0.1 \\
\quad MLP (Mean) & 1.05M & 96.1$\pm$0.0 & 84.3$\pm$0.1 & 96.7$\pm$0.0 & 87.1$\pm$0.2 & 96.8$\pm$0.2 & 87.8$\pm$0.1 & 0.52M & 96.3$\pm$0.3 & 83.9$\pm$0.3 \\
\quad TimeAttn \cite{kramar2026building} & 1.59M & 96.6$\pm$0.0 & 86.0$\pm$0.0 & 96.7$\pm$0.3 & 87.3$\pm$0.9 & 96.6$\pm$0.1 & 86.8$\pm$1.0 & 0.80M & 95.8$\pm$0.2 & 80.3$\pm$1.3 \\
\quad LSTM \cite{damirchi2026truth} & 2.57M & 96.4$\pm$0.2 & 86.3$\pm$0.1 & 97.0$\pm$0.1 & 87.6$\pm$0.2 & 96.9$\pm$0.1 & 87.8$\pm$0.2 & 1.51M & 95.9$\pm$0.2 & 81.6$\pm$0.2 \\
\midrule
\rowcolor{aliceblue} \quad  $\boldsymbol{\mathcal{D}^2}$\textbf{-MLP (Ours)} & $\leq$0.18M & \underline{96.7$\pm$0.2} & \underline{87.9$\pm$0.6} & \textbf{97.4$\pm$0.0} & \textbf{90.3$\pm$0.2} & \underline{97.3$\pm$0.0} & \underline{89.9$\pm$0.1} & 0.06M & \underline{96.9$\pm$0.1} & \underline{87.2$\pm$0.2} \\
\rowcolor{aliceblue} \quad  $\boldsymbol{\mathcal{D}^2}$\textbf{-TimeAttn (Ours)} & $\leq$0.37M & \textbf{96.9$\pm$0.0} & \textbf{88.3$\pm$0.3} & \underline{97.4$\pm$0.1} & \underline{90.1$\pm$0.2} & \textbf{97.5$\pm$0.0} & \textbf{90.4$\pm$0.1} & 0.08M & \textbf{96.9$\pm$0.1} & \textbf{87.4$\pm$0.3} \\
\bottomrule
\end{tabular}%
}
\end{table*}

\paragraph{Baselines}
We compare our method against eight baselines, organized into two categories. 
\textbf{(1) Single-step methods} use only the hidden state from the last denoising step $\mathbf{h}_{1}$ for both training and prediction. 
\textbf{(2) Multi-step methods} leverage all $S$ denoising steps. For mean-based approaches, LP and MLP are trained on the temporal mean $\bar{\mathbf{h}} = \frac{1}{S}\sum_s \mathbf{h}_s$. At test time, two prediction strategies are considered: the \textit{Mean} variant predicts directly from $\bar{\mathbf{h}}$, while the \textit{MV} variant applies the trained probe to each step independently and takes a majority vote. This yields four baselines: LP (Mean), LP (MV), MLP (Mean), and MLP (MV). 
For sequence-based approaches, TimeAttn \cite{kramar2026building} and LSTM \cite{damirchi2026truth} operate directly on the full ordered sequence $(\mathbf{h}_1, \ldots, \mathbf{h}_{S})$. TimeAttn uses a temporal attention mechanism to aggregate hidden states, while LSTM encodes the sequence with a recurrent model. Experiment details are provided in Appendix~\ref{app:exp}.


\paragraph{Evaluation Metrics}
We report three metrics. \textbf{Accuracy (Acc)} measures the fraction of correctly classified prompts. \textbf{Macro-F1} is the unweighted mean of the F1 scores for the safe and unsafe classes, each computed as the harmonic mean of that class's precision and recall. \textbf{E[P]} measures the expected number of parameters used per example at test time, capturing the effective monitor size under cascade routing, which is a proxy for runtime memory cost. For our method, $\mathrm{E}[P] = |\theta_{\text{LP}}| + \rho \cdot |\theta_{g}|$, where $|\theta_{\text{LP}}|$ and $|\theta_{g}|$ are the parameter counts of the linear probe and the advanced probe respectively, and $\rho$ is the fraction of examples routed to the advanced probe. Additional evaluation metrics, including F2-score, false rejection rate (FRR), inference time, and FLOPs, are reported in Appendix~\ref{app:more_metrics}.

\subsection{Main Results}
\cref{tab:main_results,tab:main_results_toxicchat,tab:cross_dataset} summarize the intra-dataset and cross-dataset performance. $\mathcal{D}^2${-Monitor} outperforms all baselines in both accuracy and F1 score in every model--dataset setting. 
Relative to the strongest baseline, the margin is $+1.4$ to $+3.5$ F1 intra-dataset and $+2.7$ to $+4.3$ F1 under cross-dataset transfer.
More importantly, $\mathcal{D}^2${-Monitor} achieves state-of-the-art performance while incurring substantially lower computational cost than non-linear baselines. As shown in~\cref{fig:pareto}, our method provides the best effectiveness-efficiency trade-off overall. In contrast, sequence-based baselines such as LSTM and TimeAttn incur significantly higher computational cost without delivering better performance.
We attribute the advantage of $\mathcal{D}^2${-Monitor} to its hesitation-aware routing mechanism, which accurately directs ``hard'' samples to the advanced probe for further processing~(\cref{sec:analysis}). Moreover, hesitation reflects a form of \emph{model-intrinsic} uncertainty rather than dataset-specific patterns. Consequently, the performance gains can generalize across datasets.

\begin{table}[t]
\centering
\caption{\textbf{Cross-dataset generalization.} Monitors are trained on \textit{WildGuardMix} and tested on \textit{ToxicChat} and \textit{OpenAI-Moderation}. Mean$\pm$std over seeds 0, 1, 2. Best results are in \textbf{bold}, and second-best are \underline{underlined}.}
\label{tab:cross_dataset}
\resizebox{\textwidth}{!}{%
\begin{tabular}{l c cc cc cc cc}
\toprule
& & \multicolumn{4}{c}{\textbf{LLaDA-8B-Base}} & \multicolumn{4}{c}{\textbf{LLaDA-8B-Instruct}} \\
\cmidrule(lr){3-6} \cmidrule(lr){7-10}
& & \multicolumn{2}{c}{ToxicChat} & \multicolumn{2}{c}{OpenAI-Moderation} & \multicolumn{2}{c}{ToxicChat} & \multicolumn{2}{c}{OpenAI-Moderation} \\
\cmidrule(lr){3-4} \cmidrule(lr){5-6} \cmidrule(lr){7-8} \cmidrule(lr){9-10}
\textbf{Method} & \textbf{E[P]} & Acc & F1 & Acc & F1 & Acc & F1 & Acc & F1 \\
\midrule
\textit{Single-step methods} \\[2pt]
\quad LP (Last Step) & 4$\times$10$^{-3}$M & 80.1$\pm$1.0 & 62.5$\pm$0.5 & 61.6$\pm$0.5 & 61.4$\pm$0.5 & 84.2$\pm$2.0 & 66.5$\pm$2.2 & 66.9$\pm$0.8 & 66.4$\pm$0.7 \\
\quad MLP (Last Step) & 1.05M & 82.8$\pm$1.3 & 65.1$\pm$1.3 & 61.3$\pm$1.5 & 61.2$\pm$1.4 & 85.1$\pm$2.6 & 67.8$\pm$2.6 & 65.3$\pm$2.0 & 65.0$\pm$1.9 \\
\midrule
\textit{Full-trajectory methods} \\[2pt]
\quad LP (MV) & 4$\times$10$^{-3}$M & 83.5$\pm$0.7 & 65.8$\pm$1.3 & 59.6$\pm$3.6 & 59.5$\pm$3.5 & 84.6$\pm$0.7 & 67.1$\pm$0.7 & 63.8$\pm$0.8 & 63.5$\pm$0.8 \\
\quad LP (Mean) & 4$\times$10$^{-3}$M & 84.2$\pm$1.0 & 66.5$\pm$1.4 & 60.0$\pm$3.6 & 59.9$\pm$3.5 & 85.3$\pm$0.4 & 67.8$\pm$0.5 & 64.3$\pm$1.0 & 63.9$\pm$1.0 \\
\quad MLP (MV) & 1.05M & 79.4$\pm$1.7 & 62.3$\pm$1.3 & 60.9$\pm$2.2 & 60.8$\pm$2.1 & 86.9$\pm$1.1 & 70.5$\pm$1.3 & 62.4$\pm$0.7 & 62.2$\pm$0.6 \\
\quad MLP (Mean) & 1.05M & 82.5$\pm$1.5 & 65.4$\pm$1.3 & 62.2$\pm$2.1 & 62.1$\pm$2.0 & 87.3$\pm$1.0 & 71.1$\pm$1.1 & 63.0$\pm$0.8 & 62.7$\pm$0.7 \\
\quad TimeAttn \cite{kramar2026building} & 1.59M & 84.3$\pm$1.9 & 67.7$\pm$1.5 & 66.0$\pm$2.1 & 65.4$\pm$1.9 & 88.2$\pm$0.8 & 72.2$\pm$1.1 & 67.5$\pm$1.9 & 67.1$\pm$1.8 \\
\quad LSTM \cite{damirchi2026truth} & 2.57M & 83.2$\pm$2.0 & 66.1$\pm$3.0 & 66.1$\pm$2.3 & 65.5$\pm$2.0 & 88.1$\pm$1.8 & 72.2$\pm$2.1 & 68.6$\pm$2.8 & 67.8$\pm$2.2 \\
\midrule
\rowcolor{aliceblue} \quad $\boldsymbol{\mathcal{D}^2}$\textbf{-MLP (Ours)} & $\leq$0.68M & \textbf{88.4$\pm$0.7} & \textbf{72.0$\pm$0.2} & \underline{67.2$\pm$0.8} & \underline{66.8$\pm$0.7} & \underline{89.2$\pm$0.2} & \underline{73.2$\pm$0.1} & \underline{69.6$\pm$0.7} & \underline{68.9$\pm$0.6} \\
\rowcolor{aliceblue} \quad $\boldsymbol{\mathcal{D}^2}$\textbf{-TimeAttn (Ours)} & $\leq$0.93M & \underline{87.7$\pm$0.4} & \underline{71.6$\pm$0.8} & \textbf{69.4$\pm$1.1} & \textbf{68.6$\pm$0.8} & \textbf{90.6$\pm$0.6} & \textbf{75.3$\pm$0.7} & \textbf{71.3$\pm$0.7} & \textbf{70.5$\pm$0.7} \\
\bottomrule
\end{tabular}%
}
\end{table}

\subsection{Analysis}
\label{sec:analysis}
\paragraph{Efficiency-Effectiveness Tradeoff}
Model providers can run safety probes alongside the D-LLM with negligible
overhead, but user-side deployment operates under a tighter budget. Raising
$\lambda$ lowers the expected parameter count $\mathrm{E}[P]$, which reflects
runtime memory cost, at the risk of missing difficult cases. In practice the
trade-off is mild: sweeping $\lambda$ at test time without retraining changes
F1 by less than 0.8 points on ToxicChat (Fig.~\ref{fig:mdt_sensitivity_indomain}). As
shown in Fig.~\ref{fig:pareto}, $\mathcal{D}^2$-Monitor achieves the best F1 while using fewer parameters than most baselines. 

\paragraph{Comparison with LLM-Based Monitors}
To situate $\mathcal{D}^2$-Monitor against the LLMs-as-monitors paradigm, we fine-tune Llama-Guard-3-1B \cite{metallamaguard3}
separately on WildGuardMix and ToxicChat and evaluate it on each.
It attains 89.9 and 90.4 F1 respectively~(\cref{tab:llamaguard3-1b-align}).
$\mathcal{D}^2$-Monitor achieves comparable performance on most models using fewer than
0.05\% parameters. Under transfer the comparison favors the LLM monitor: aligned on
WildGuardMix, it reaches 80.4 F1 on ToxicChat and 74.5 on OpenAI-Moderation,
against $71.6$--$75.3$ and $66.8$--$70.5$ for $\mathcal{D}^2$-Monitor~(\cref{tab:llm_monitor_comparison}). Note that every probe-based method in
\cref{tab:cross_dataset} degrades sharply under distribution shift, so this is not specific to our design. We regard this as an open problem for
representation-based monitors~\cite{wang2026false,orgad2025llms}.

\paragraph{Robustness to Generation Length and Step Length} In practice, D-LLMs can be deployed with varying generation lengths and step lengths. Training a separate monitor for each configuration is costly and impractical. A desirable property is therefore to train once under a single configuration and generalize to others. To evaluate this, we train all methods with generation length 128 and step length 4, and test under varying settings without retraining. We vary the step length $L_S \in \{1, 2, 4, 8\}$ with generation length fixed at 128 (\cref{fig:robustness_steps}), and the generation length $L \in \{16, 32, 64, 128\}$ with step length fixed at 1 (\cref{fig:robustness_length}). $\mathcal{D}^2$-Monitor consistently outperforms all baselines across both axes of variation, confirming that our method transfers reliably across decoding configurations.

\begin{figure}[t]
    \centering
    \begin{minipage}[b]{0.32\textwidth}
        \centering
        \includegraphics[width=\textwidth]{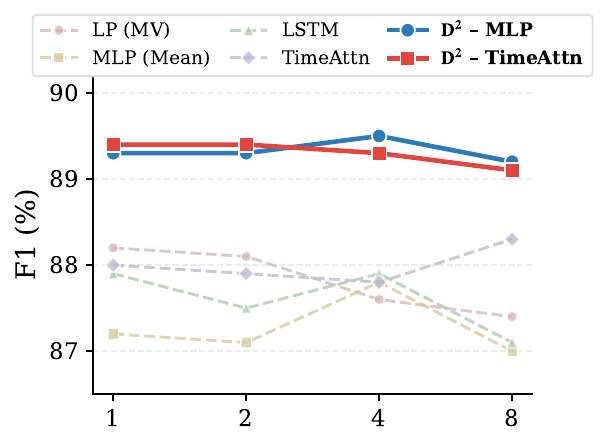}
        \subcaption{}
        \label{fig:robustness_steps}
    \end{minipage}
    \hfill
    \begin{minipage}[b]{0.32\textwidth}
        \centering
        \includegraphics[width=\textwidth]{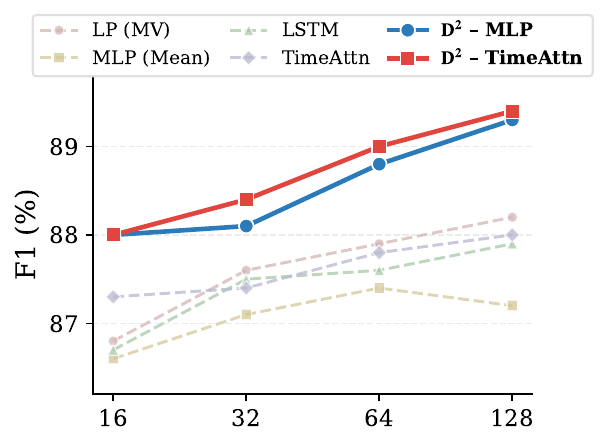}
        \subcaption{}
        \label{fig:robustness_length}
    \end{minipage}
    \hfill
    \begin{minipage}[b]{0.32\textwidth}
        \centering
        \includegraphics[width=\textwidth]{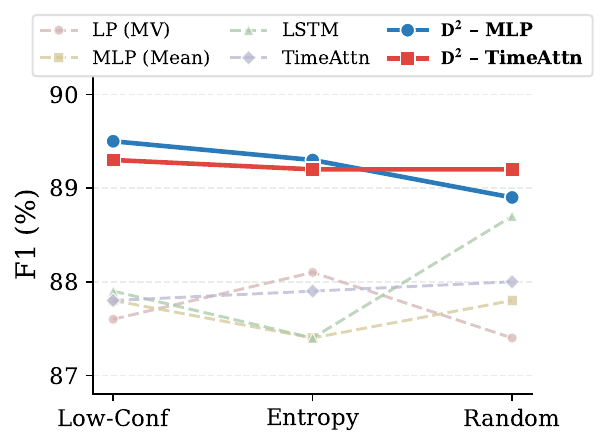}
        \subcaption{}
        \label{fig:remasking}
    \end{minipage}
    \caption{
    \textbf{(a)} Performance with different step lengths with generation length fixed at 128.
    \textbf{(b)} Performance with different generation lengths with step length fixed at 1. 
    \textbf{(c)} Performance under different remasking strategies. 
    All results are reported as F1 using LLaDA-8B-Instruct on WildGuardMix.
    }
    \label{fig:robustness_all}
\end{figure}


\paragraph{Robustness to Remasking Strategy}
We evaluate the impact of different remasking strategies on detection performance. All methods are trained under low-confidence remasking and tested under three strategies: low-confidence, entropy, and random. As shown in~\cref{fig:remasking}, $\mathcal{D}^2$-Monitor maintains consistent superiority across all strategies. 



\begin{figure*}[t]
\centering

\noindent
\begin{minipage}[t]{0.66\linewidth}
\vspace{0pt}
\centering
\begin{subfigure}[t]{0.48\linewidth}
    \centering
    \includegraphics[width=\linewidth]{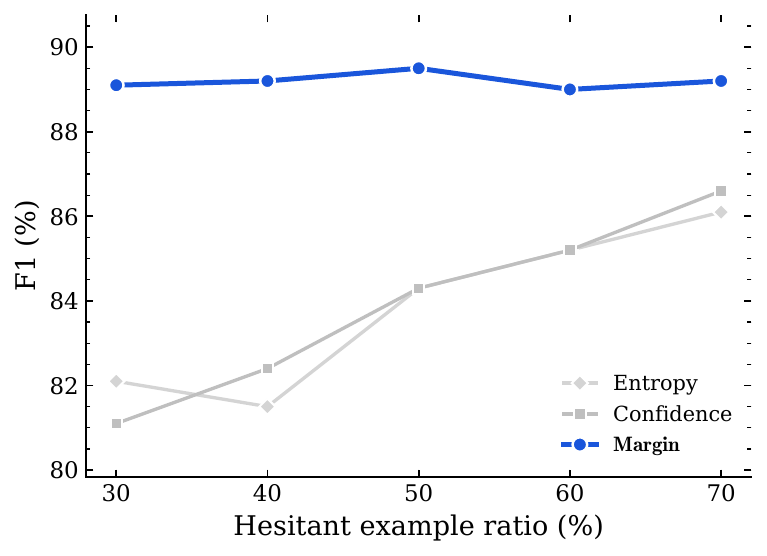}
    \caption{$\mathit{D^2}$--MLP}
\end{subfigure}
\hfill
\begin{subfigure}[t]{0.48\linewidth}
    \centering
    \includegraphics[width=\linewidth]{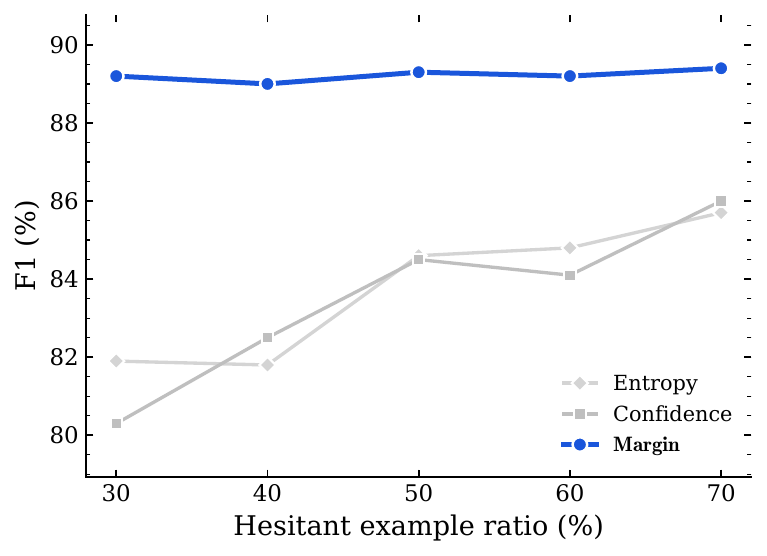}
    \caption{$\mathit{D^2}$--TimeAttn}
\end{subfigure}
\end{minipage}%
\hfill
\begin{minipage}[t]{0.30\linewidth}
\vspace{10pt}
\small
\addtocounter{figure}{-1}%
\refstepcounter{figure}%
Figure \thefigure:
Ablation on routing signals under different hesitant ratios. Margin-based routing consistently outperforms entropy and confidence, and shows greater robustness to threshold selection.
\label{fig:ablation}
\end{minipage}
\end{figure*}

\paragraph{Ablation on Routing Signal}
We replace the margin-based criterion with entropy- and confidence-based ones
on ToxicChat, retraining the advanced probe on each signal's
hesitation subset and holding the rest of the pipeline fixed. Across both
$\mathcal{D}^2$-MLP and $\mathcal{D}^2$-TimeAttn at hesitation ratios from
30\% to 70\%, the margin signal gives the highest and most stable F1~(\cref{fig:ablation}),
indicating that probe-intrinsic uncertainty routes better than probe-extrinsic
uncertainty. 
At a $50\%$
hesitation ratio, the margin signal produces the largest accuracy gap between
the base and advanced probes on the routed subset ($13.5\%$ / $12.4\%$), well
above entropy ($7.2\%$ / $7.4\%$) and confidence ($7.1\%$ / $7.3\%$), so it
isolates the samples that gain most from escalation. That subset is
also disproportionately adversarial: $64$--$88\%$ on \textit{WildGuardMix}
against a dataset-wide baseline of $47\%$~(\cref{app:hesitation_adv}), so
margin-based routing concentrates the advanced probe on likely attacks rather
than on generically hard inputs.

\vspace{0.5em}
\paragraph{Effect of Curation and Routing}
We ablate the two ingredients separately on LLaDA-8B-Instruct~(\cref{tab:ablation_2x2}):
curation, which trains the advanced probe on hesitation windows, and routing,
which applies it only to escalated samples. Both help on their own, and combining
them brings a further improvement. Averaged over the two datasets, routing alone
adds $+0.9/{+}1.0$ F1 for MLP/TimeAttn, curation $+1.0/{+}0.5$, and the
combination $+2.5/{+}2.2$. Routing also cuts $\mathrm{E}[P]$ by $4$--$7\times$.
In every case, curation on top of routing is worth $+1.0$ to $+2.0$ F1 at unchanged
$\mathrm{E}[P]$, and routing on top of curation is $4$--$7\times$ cheaper and
more accurate.

\vspace{-0.5em}
\section{Conclusion}
We showed that the denoising trajectory of a D-LLM carries a usable difficulty
signal: on some inputs the intermediate states repeatedly fall within a small
margin of a linear probe's decision boundary. The number of these hesitation
steps $n_\tau$ predicts when the probe will fail. Acting on this signal, $\mathcal{D}^2$-Monitor escalates samples whose
$n_\tau$ exceeds a threshold to a more expressive probe that reads only the
hesitation window. On three safety datasets across four
D-LLMs, it achieves state-of-the-art performance both in-domain and
cross-dataset, with the best
effectiveness--efficiency trade-off among eight baselines and at most 0.93M
expected parameters. 

\vspace{0.5em}
\paragraph{Limitations}
Although we expect these results to extend to larger
models \cite{bie2025llada2, bie2026llada2p1}, our experiments are currently
limited to D-LLMs of up to 16B parameters due to computational constraints.
Our method is stable across generation configurations without retraining
(Fig.~\ref{fig:robustness_all}), but calibration under shifts in checkpoint or safety policy remains an open problem \cite{ovadia2019can}. 
Beyond these empirical gaps,
a theoretical account of why hesitation arises during denoising would also be
interesting. Appendix~\ref{app:limit} expands on these limitations.

\newpage

\subsection*{AI use statement}
In this work, we used generative AI tools to provide feedback on research
methodology and experiments, to assist in implementing methods, to assist with
translation, and to help interpret results. We have not used generative AI tools
to generate synthetic data sets, to develop the theoretical models or conceptual
framework of this work, to propose or refine hypotheses, to clean or reformat
datasets, or to support qualitative and thematic data analysis; formulating
mathematical claims, providing critical ingredients for proving mathematical
claims, and assisting in the writing of proofs are not applicable to this work.
Additionally, we used generative AI tools to edit this paper
to improve readability, to suggest a structure for the paper, and to create or edit software code.
All AI-assisted work was reviewed by the authors: AI-assisted code was read and
tested for correctness, and AI-assisted
text was revised before inclusion. We take responsibility for the final content of this work, including
text, claims or artifacts produced with the aid of generative AI.

\subsection*{Ethics statement}
This work involves no human subjects and releases no new data or models; all
experiments use publicly available safety benchmarks under their respective
licenses (Appendix~\ref{app:broader-impacts}). Two risks are worth noting. First, like any safety classifier, our monitor can falsely block benign prompts;
we report false rejection rates alongside detection performance
(Appendix~\ref{app:more_metrics}), where $\mathcal{D}^2$-Monitor attains the lowest rate
among all methods on the four D-LLMs we evaluate. Second, because routing depends on hesitation severity, an adversary
who crafts inputs that keep hesitation low could remain on the lightweight path
and evade the advanced probe. 

\subsection*{Reproducibility statement}
All reported numbers are means over three random seeds (0, 1, and 2), reported
with standard deviations. The datasets, splits, models, and evaluation protocol
are specified in Appendix~\ref{app:exp}; probe architectures and their exact dimensions
in Appendix~\ref{app:probe_architectures}; and the procedure for selecting the test-time
hyperparameters $\tau$ and $\lambda$ in Appendix~\ref{sec:appendix_hparam}. 


\bibliographystyle{iclr2027_conference}
\bibliography{references.bib}

\newpage
\appendix
\renewcommand{\floatpagefraction}{0.45}
\renewcommand{\bottomfraction}{0.50}
\setcounter{topnumber}{2}
\setcounter{bottomnumber}{1}
\setcounter{totalnumber}{2}

\appendix

\section{Limitation}
\label{app:limit}

\paragraph{Model Scale} Our experiments cover four D-LLMs, from 8B to 16B
parameters. Within this range the advantage of $\mathcal{D}^2$-Monitor does not
shrink with model size. On LLaDA-2.0-mini, the largest model we evaluate, it
improves F1 over the strongest baseline by $2.7$ points on \textit{WildGuardMix}
and $3.5$ points on \textit{ToxicChat}, compared with $1.4$ to $1.9$ and $2.0$
to $2.7$ points on the 8B models. We therefore expect the approach to carry over
to larger D-LLMs \cite{bie2025llada2, bie2026llada2p1}. Absolute F1 is lowest on
this model for every method, which may be due to its narrower hidden states.
\paragraph{Calibration under Distribution Shift} Calibration under distribution
shift is a long-standing open problem \cite{ovadia2019can, koh2021wilds}, and
we do not claim to solve it. A deployed D-LLM may run with a different
generation length, step length, or remasking strategy, and its checkpoint and
safety policy may change over time. A probe-based monitor, in contrast, is
fitted once on representations from a single setting. However, two properties of
$\mathcal{D}^2$-Monitor partially mitigate this. A probe trained under one
generation configuration transfers to others without retraining
(Figure~\ref{fig:robustness_all}), and performance is flat over a wide range of
$\lambda$ (Appendix~\ref{app:sensitivity}), so the routing threshold does not
have to be re-tuned precisely.

\paragraph{Why Hesitation Arises} We characterize hesitation empirically and
show that it predicts when the base probe fails, but we do not explain why the
intermediate states of some inputs repeatedly fall close to the decision
boundary. Work on the internals of D-LLMs is still early. Recent studies
describe how representations are organized across layers and denoising steps
\cite{goel2026skip} and decompose them with sparse autoencoders
\cite{wang2026dlm}, but neither accounts for the dynamics of the probe margin
along a trajectory. Why some trajectories hesitate, and whether this can be
predicted before decoding starts, remains open. Answering either would put the
routing rule on firmer ground, and we see it as an interesting direction for
future work.
\section{Broader Impacts, Safeguards, and Licenses}
\label{app:broader-impacts}
\paragraph{Broader Impacts}
Our work has clear positive societal impact: $\mathcal{D}^2$-Monitor provides a lightweight, always-on safety mechanism for D-LLMs that can detect harmful or adversarial inputs at low computational cost, which is particularly valuable for resource-constrained or edge deployment where running heavyweight LLM-as-monitor solutions is infeasible. By improving the practicality of safety monitoring, our work helps reduce the risk of D-LLMs being misused for harmful content generation.
We also acknowledge potential negative impacts. (1) Like any safety classifier, the monitor may incur false rejections on benign inputs, potentially over-blocking legitimate queries; we report FRR explicitly in Appendix~\ref{app:more_metrics} to make this trade-off transparent. (2) As discussed in Appendix~\ref{app:limit}, hesitation-based routing could in principle be targeted by adaptive adversaries who craft prompts that suppress hesitation steps to evade the second-stage probe. We recommend that practitioners deploying $\mathcal{D}^2$-Monitor pair it with complementary defenses and continue to monitor for adversarial drift over time.

\paragraph{Safeguards}
The paper does not release any high-risk pretrained models, generative models, or scraped datasets. The artifacts produced by our work are lightweight safety probes ($\leq 0.93$M parameters) trained on publicly available safety benchmarks; these probes are themselves a defensive mechanism intended to mitigate misuse rather than enable it. The base D-LLMs (LLaDA family) and safety datasets we build on are released by their original authors under their respective licenses, which we do not modify or redistribute.

\paragraph{Licenses for Existing Assets}
All existing assets used in this work are properly cited and used in accordance with their licenses.

\textbf{Datasets} WildGuardMix~\cite{han2024wildguard} is released under the ODC-BY license; ToxicChat~\cite{lin2023toxicchat} is released under CC-BY-NC-4.0 and used for non-commercial research only; OpenAI-Moderation~\cite{openai2022moderation} is released under the MIT license. 

\textbf{Models} LLaDA-8B-Base, LLaDA-8B-Instruct~\cite{nie2025large}, and LLaDA-1.5~\cite{zhu2025llada} are released under the MIT license; LLaDA-2.0-mini~\cite{bie2025llada2} is released under the Apache-2.0 license. 

\section{Experiment Details}
\label{app:exp}
\subsection{Hyperparameter Tuning}
\label{sec:appendix_hparam}

We tune all hyperparameters on a validation set derived from the training data. 
Specifically, we split the original training set into training and validation subsets with a ratio of 4:1.
To ensure fair comparison across different probe architectures, we control model capacity 
by fixing key architectural dimensions (e.g., hidden size or hidden dimension) for each probe, 
rather than tuning them extensively. For each probe, we define the hyperparameter search space as follows:

\paragraph{LinearProbe:}
\begin{itemize}
    \item \textbf{Learning rate:} $\{3\mathrm{e}{-3}, 1\mathrm{e}{-3}, 3\mathrm{e}{-4}, 1\mathrm{e}{-4}, 3\mathrm{e}{-5}, 1\mathrm{e}{-5}\}$
    \item \textbf{Weight decay:} $\{0, 1\mathrm{e}{-6}, 1\mathrm{e}{-5}, 1\mathrm{e}{-4}, 1\mathrm{e}{-3}\}$
\end{itemize}
\textbf{MLP:}
\begin{itemize}
    \item \textbf{Learning rate:} $\{1\mathrm{e}{-3}, 3\mathrm{e}{-4}, 1\mathrm{e}{-4}, 3\mathrm{e}{-5}, 1\mathrm{e}{-5}\}$
    \item \textbf{Weight decay:} $\{0, 1\mathrm{e}{-6}, 1\mathrm{e}{-5}, 1\mathrm{e}{-4}\}$
    \item \textbf{Dropout:} $\{0, 0.1, 0.2, 0.3\}$
    \item \textbf{Hidden dimension ($K$):} $256$
\end{itemize}
\textbf{TimeAttn:}
\begin{itemize}
    \item \textbf{Learning rate:} $\{1\mathrm{e}{-3}, 3\mathrm{e}{-4}, 1\mathrm{e}{-4}, 3\mathrm{e}{-5}, 1\mathrm{e}{-5}\}$
    \item \textbf{Weight decay:} $\{0, 1\mathrm{e}{-6}, 1\mathrm{e}{-5}, 1\mathrm{e}{-4}\}$
    \item \textbf{Dropout:} $\{0, 0.1, 0.2, 0.3\}$
    \item \textbf{MLP hidden dimension ($K$):} $256$
    \item \textbf{Attention dimension ($d_a$):} $128$
\end{itemize}
\textbf{LSTM:}
\begin{itemize}
    \item \textbf{Projection dimension ($d_p$):} $512$
    \item \textbf{Hidden size ($d_h$):} $128$
    \item \textbf{Learning rate:} $\{1\mathrm{e}{-5}, 1\mathrm{e}{-4}, 1\mathrm{e}{-3}\}$
    \item \textbf{Weight decay:} $\{0, 1\mathrm{e}{-6}, 1\mathrm{e}{-5}, 1\mathrm{e}{-4}\}$
    \item \textbf{Dropout:} $\{0, 0.1, 0.2, 0.3\}$
\end{itemize}

\paragraph{Baseline Methods}

For all baseline methods, hyperparameters are selected via grid search on the validation set 
using the search ranges above. The best configuration is chosen based on validation performance, 
and the model is retrained on the full training set before evaluation on the test set.

\paragraph{Our Method}
$\mathcal{D}^2$-Monitor has two trained components whose hyperparameters cannot be
selected independently. The linear probe used for out-of-fold (OOF) scoring on the
training set determines which steps are labelled as hesitation, and therefore which
examples the advanced probe is trained on; the advanced probe in turn determines
final performance. We treat the two stages as a joint search: for each candidate
configuration of the OOF probe we train the corresponding advanced probe (MLP or
TimeAttn), and select the pair by validation performance. 

\paragraph{Fairness and Evaluation Protocol}
All hyperparameter tuning is performed strictly on the validation set, and no test
data is used at any stage of model selection.

\paragraph{Test-Time Hyperparameters}
Our method involves two test-time hyperparameters: the hesitation threshold $\tau$
and the routing parameter $\lambda$. Neither requires retraining, so both can be
chosen by evaluating the trained monitor on the validation set.

\begin{figure*}[p]
\centering
\newcommand{\subfw}{0.48\textwidth}

\begin{subfigure}[t]{\subfw}
    \centering
    \includegraphics[width=\linewidth]{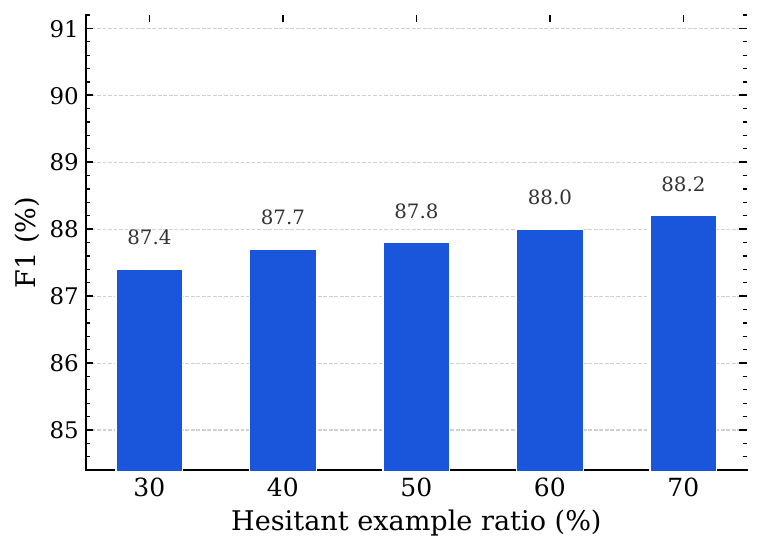}
    \caption{LLaDA-8B-Base: $D^2$-MLP}
\end{subfigure}
\hfill
\begin{subfigure}[t]{\subfw}
    \centering
    \includegraphics[width=\linewidth]{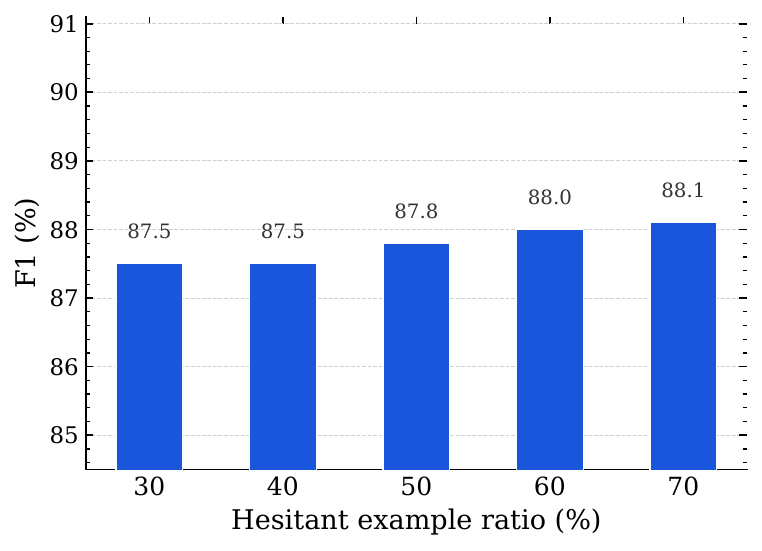}
    \caption{LLaDA-8B-Base: $D^2$-TimeAttn}
\end{subfigure}

\vspace{2pt}

\begin{subfigure}[t]{\subfw}
    \centering
    \includegraphics[width=\linewidth]{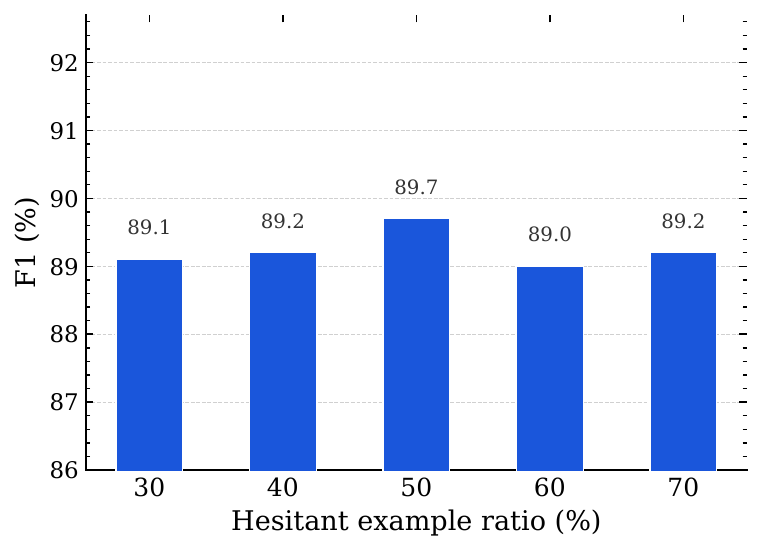}
    \caption{LLaDA-8B-Instruct: $D^2$-MLP}
\end{subfigure}
\hfill
\begin{subfigure}[t]{\subfw}
    \centering
    \includegraphics[width=\linewidth]{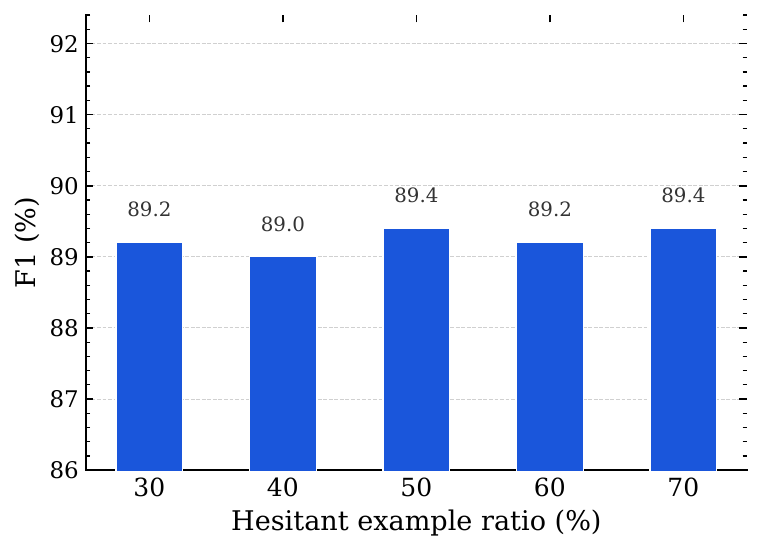}
    \caption{LLaDA-8B-Instruct: $D^2$-TimeAttn}
\end{subfigure}

\vspace{2pt}

\begin{subfigure}[t]{\subfw}
    \centering
    \includegraphics[width=\linewidth]{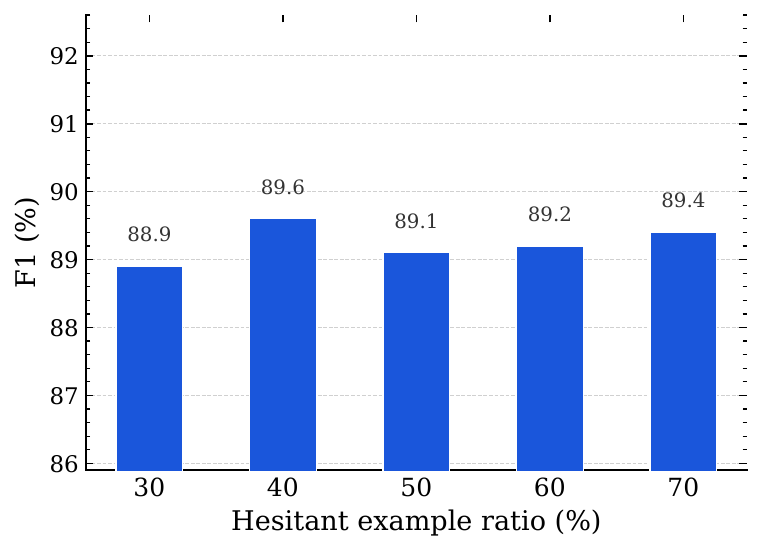}
    \caption{LLaDA-1.5: $D^2$-MLP}
\end{subfigure}
\hfill
\begin{subfigure}[t]{\subfw}
    \centering
    \includegraphics[width=\linewidth]{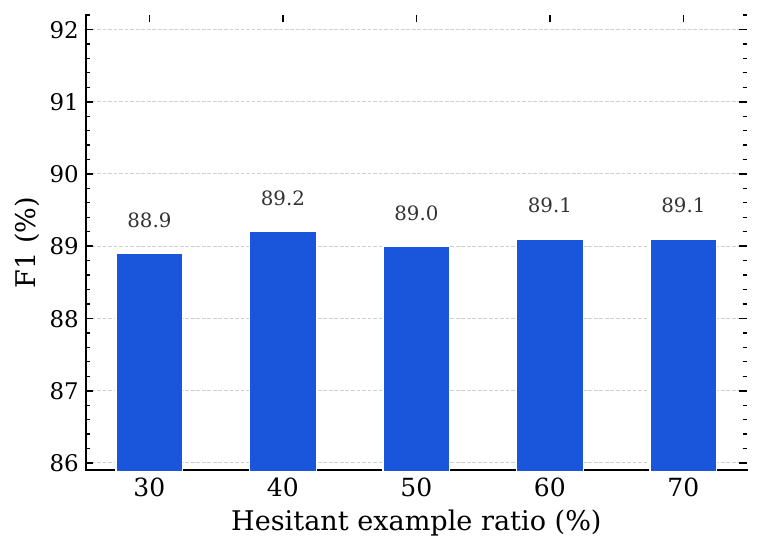}
    \caption{LLaDA-1.5: $D^2$-TimeAttn}
\end{subfigure}

\vspace{2pt}

\begin{subfigure}[t]{\subfw}
    \centering
    \includegraphics[width=\linewidth]{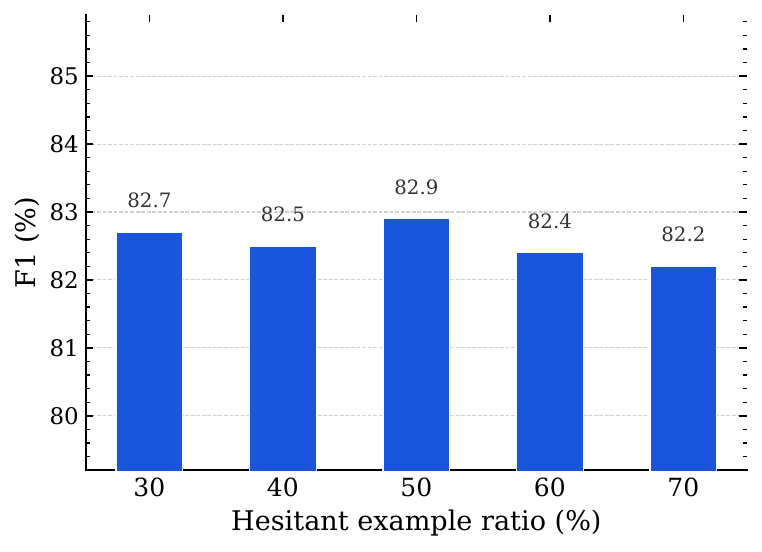}
    \caption{LLaDA-2.0-mini: $D^2$-MLP}
\end{subfigure}
\hfill
\begin{subfigure}[t]{\subfw}
    \centering
    \includegraphics[width=\linewidth]{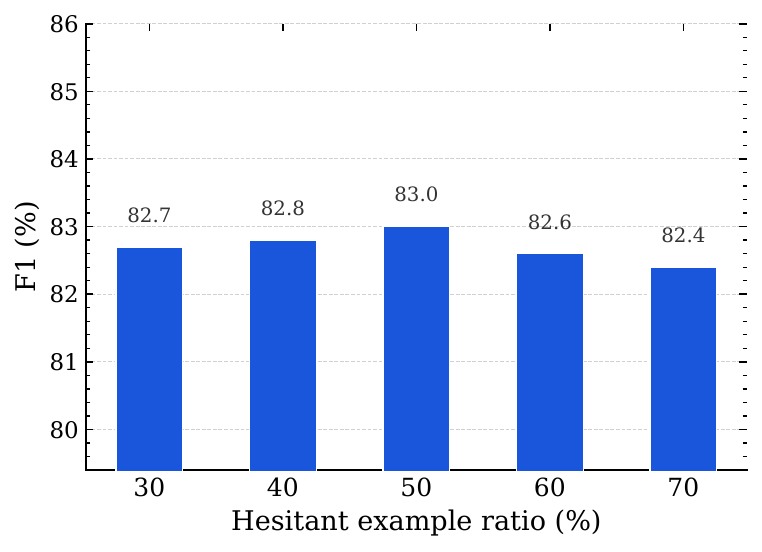}
    \caption{LLaDA-2.0-mini: $D^2$-TimeAttn}
\end{subfigure}

\caption{
F1 score (\%) vs.\ hesitant example ratio under $D^2$-MLP (left) and $D^2$-TimeAttn (right) across four LLaDA variants. All probes are trained and tested on \textit{WildGuardMix}.
}
\label{fig:multi_model_results}
\end{figure*}

\textbf{Choice of $\tau$}
The threshold $\tau$ controls a trade-off between coverage and purity of hesitation
examples. A larger $\tau$ includes more training samples but may introduce stable
(non-hesitant) steps, while a smaller $\tau$ yields cleaner hesitation signals at the
cost of fewer training examples. To balance this trade-off, we select $\tau$ such
that approximately 50\% of the samples are identified as hesitation examples. As
shown in Fig.~\ref{fig:multi_model_results}, varying this proportion from 30\% to
70\% leads to only minor performance differences, indicating that our method is not
sensitive to the exact choice of $\tau$. For intra-dataset evaluation we fix $\tau$
at the value selected on the training data and reuse it at test time, since the
data distributions are similar.

\textbf{Selecting the operating point}
Given $\tau$, we select $\lambda$ in two steps. We first locate the configuration
with the highest validation F1; we then restrict attention to all configurations
whose validation F1 fall slightly below that maximum, and among these take the one
with the smallest expected parameter count $\mathrm{E}[P]$. This selects the
cheapest point on the validation plateau rather than the nominally best one,
trading no measurable detection performance for lower test-time cost. The resulting
operating point is then applied unchanged at test time.

\textbf{Cross-dataset evaluation}
For cross-dataset transfer (trained on WildGuardMix and evaluated on ToxicChat
and OpenAI-Moderation), the distribution shift means $\tau$ can no longer be
carried over from training. We therefore run a grid search over $(\tau,\lambda)$
jointly on a held-out validation split of the target dataset. The operating
point is then chosen from this grid by the same rule as before: among all
configurations whose validation F1 fall slightly below that maximum, we take
the one with the smallest $\mathrm{E}[P]$. As shown in
\cref{fig:tau_mdt_sensitivity_cross}, F1 varies little across the entire grid, so
the reported results do not hinge on which configuration the search happens to
return.

\subsection{Probe Architectures}
\label{app:probe_architectures}

Here we provide specific details about the architectures of all probe baselines considered.
Given the step-level trajectory $\mathbf{H}=[\mathbf{h}_1,\ldots,\mathbf{h}_S]\in \mathbb{R}^{D \times S}$, each probe instantiates a particular $f\in\mathcal{F}$ that maps $\mathbf{H}$ to a scalar logit $s\in\mathbb{R}$ for binary classification.
When presenting architectures below, we omit bias terms in the hidden layer(s) and output for brevity.
All probes are trained with Adam with a batch size of 256 for 50 epochs on an NVIDIA A40 GPU with 48GB VRAM.
Each individual probe performs its own grid search over the hyperparameters listed in Appendix~\ref{sec:appendix_hparam}.

\paragraph{Normalization}
Before training, we normalize the activations using statistics computed from the training split. We use two normalization modes depending on the probe type. For \emph{full-trajectory methods}, we apply \emph{per-feature} normalization: the mean and standard deviation are computed over both the sample and step axes, yielding statistics of shape $[D]$. For \emph{single-step methods}, we apply \emph{per-step} normalization: statistics are computed per denoising step, yielding shape $[S, D]$.

\paragraph{Pooling Strategies}
For probes that require a fixed-length input (Linear Probe and MLP), we reduce the step dimension via one of three pooling strategies applied after normalization:
\begin{itemize}
    \item \textbf{Mean pooling:} $\bar{\mathbf{h}} = \frac{1}{S}\sum_{s=1}^{S} \mathbf{h}_s \in \mathbb{R}^D$, used at both train and test time.
    \item \textbf{Last-step:} $\bar{\mathbf{h}} = \mathbf{h}_1 \in \mathbb{R}^D$, using only the final step's activation at both train and test time. Normalization statistics are computed from the last step only.
    \item \textbf{Majority vote (MV):} The probe is trained on mean-pooled features. At test time, each step $\mathbf{h}_s$ is classified independently, and the final prediction is determined by majority vote across $S$ steps: $\hat{y} = \mathbf{1}\!\left[\sum_{s=1}^{S} \mathbf{1}[\hat{y}_s = 1] \geq S/2\right]$.
\end{itemize}

\paragraph{Linear Probe (LP)}
After pooling, the linear probe computes:
\begin{equation}
    s = \mathbf{w}^\top \bar{\mathbf{h}}\,,
\end{equation}
with $\mathbf{w} \in \mathbb{R}^{D}$.
Combined with the three pooling strategies, this yields three variants: LP~(Mean), LP~(Last Step), and LP~(MV).

\paragraph{MLP}
The two-layer MLP probe computes:
\begin{equation}
    s = \mathbf{W}_{\mathrm{out}}\, \mathrm{Dropout}\!\left(\mathrm{ReLU}\!\left(\mathbf{W}_{\mathrm{in}} \bar{\mathbf{h}}\right)\right),
\end{equation}
with $\mathbf{W}_{\mathrm{in}} \in \mathbb{R}^{K \times D}$ and $\mathbf{W}_{\mathrm{out}} \in \mathbb{R}^{1 \times K}$, where $K = 256$ is the hidden dimension.
Analogously to the linear probe, this yields three variants: MLP~(Mean), MLP~(Last Step), and MLP~(MV).

\paragraph{TimeAttn}
This probe operates directly on the full trajectory $\mathbf{H} \in \mathbb{R}^{D \times S}$ without step-level pooling.
It first applies layer normalization, then computes attention weights over denoising steps via an additive attention mechanism:
\begin{equation}
    \alpha_s = \frac{\exp\!\left(\mathbf{v}^\top \tanh\!\left(\mathbf{W}_a \mathbf{h}_s\right)\right)}{\sum_{s'=1}^{S} \exp\!\left(\mathbf{v}^\top \tanh\!\left(\mathbf{W}_a \mathbf{h}_{s'}\right)\right)}\,,
\end{equation}
where $\mathbf{W}_a \in \mathbb{R}^{d_a \times D}$ and $\mathbf{v} \in \mathbb{R}^{d_a}$ with attention dimension $d_a = 128$.
The attended representation $\mathbf{c} = \sum_{s=1}^{S} \alpha_s \mathbf{h}_s \in \mathbb{R}^D$ is then classified via a two-layer MLP with layer normalization:
\begin{equation}
    s = \mathbf{W}_2\, \mathrm{Dropout}\!\left(\mathrm{ReLU}\!\left(\mathbf{W}_1\, \mathrm{LN}(\mathbf{c})\right)\right),
\end{equation}
with $\mathbf{W}_1 \in \mathbb{R}^{K \times D}$, $\mathbf{W}_2 \in \mathbb{R}^{1 \times K}$, and $K = 256$.

\paragraph{LSTM}
The LSTM probe first projects each step representation through a feedforward layer with layer normalization:
\begin{equation}
    \mathbf{h}'_s = \mathrm{GELU}\!\left(\mathbf{W}_{\mathrm{proj}}\, \mathrm{LN}(\mathbf{h}_s)\right),
\end{equation}
with $\mathbf{W}_{\mathrm{proj}} \in \mathbb{R}^{d_p \times D}$ and projection dimension $d_p = 512$.
The projected sequence is then processed by a 2-layer unidirectional LSTM:
\begin{equation}
    \tilde{\mathbf{h}}_1, \ldots, \tilde{\mathbf{h}}_S = \mathrm{LSTM}\!\left(\mathbf{h}'_1, \ldots, \mathbf{h}'_S\right),
\end{equation}
with hidden size $d_h = 128$.
The final hidden state is classified via a layer-normalized linear head:
\begin{equation}
    s = \mathbf{w}_{\mathrm{out}}^\top \mathrm{LN}(\tilde{\mathbf{h}}_S)\,,
\end{equation}
with $\mathbf{w}_{\mathrm{out}} \in \mathbb{R}^{d_h}$.

\section{Additional Results}
\subsection{More Evaluation Metrics}
\label{app:more_metrics}

\paragraph{Additional Metrics}
Beyond accuracy and macro-F1 reported in the main text, we evaluate four additional metrics. The \textbf{F2-score} is an instance of the F$_\beta$-score with $\beta=2$, which weights recall $\beta^2$ times as much as precision, reflecting a preference in safety monitoring for catching harmful content at the cost of slightly more false alarms. \textbf{Precision} measures the fraction of inputs flagged as unsafe that are truly unsafe; high precision means the monitor rarely triggers on benign content, reducing unnecessary refusals. \textbf{Recall} measures the fraction of truly unsafe inputs correctly identified; high recall ensures that harmful prompts are not missed in safety-sensitive deployments. Finally, the \textbf{False Rejection Rate (FRR)} is the fraction of benign inputs incorrectly flagged as unsafe, i.e., $\mathrm{FRR} = \mathrm{FP} / (\mathrm{FP} + \mathrm{TN})$; unlike $1 - \mathrm{Precision}$, which conditions on positive predictions, FRR conditions on the actual benign population and thus directly measures how often legitimate users are unnecessarily blocked (lower is better).

Tables~\ref{tab:wgm_llada_8b_base}--\ref{tab:wgm_llada_20_mini} report all six metrics for each model. The results are consistent with the main findings: $\mathcal{D}^2$-MLP and $\mathcal{D}^2$-TimeAttn achieve the best or second-best performance across nearly all metrics and models, confirming that the gains observed in accuracy and F1 are not obtained at the expense of precision, recall balance, or false refusal rate.

\paragraph{Inference Time}
We report the inference time of different methods on WildGuardMix with 1,725 prompts. 
To isolate the overhead of the safety monitor, timing starts after the base model has completed generation and all hidden states have been extracted.
For LP (Last Step), LP (MV), MLP (Last Step), and MLP (MV), 
we measure only the time required to process the last step. 
This is because, in typical deployment, intermediate steps are processed online during generation 
(i.e., one step is processed as it is generated), and thus do not introduce additional latency after generation.
For LP (Mean) and MLP (Mean), the reported time includes both the cost of computing the mean over all hidden states and the subsequent probe evaluation.
In contrast, TimeAttn and LSTM operate on the full trajectory after generation, 
and their inference time includes processing all steps.
For our method, the total inference time consists of two components: 
(i) the linear probe applied to the last step, and 
(ii) the additional computation of the advanced probe (e.g., MLP or TimeAttn) over the selected window when routing is triggered.

As shown in \cref{tab:inference_time_all}, D$^2$-Monitor substantially reduces
inference cost relative to probes that score and aggregate every denoising
step. Compared to MLP (Mean), D$^2$-MLP is 2.4$\times$--6.6$\times$ faster
across the four LLaDA models, and D$^2$-TimeAttn is 4.1$\times$--5.1$\times$
faster than TimeAttn. D$^2$-MLP also stays within the same order of magnitude
as single-step probes such as LP (Last Step) and MLP (Last Step), indicating
that the routing logic itself adds negligible overhead.

\paragraph{FLOPs Computation}
We analytically compute the expected FLOPs per sample for each probe based on the architectures detailed in Appendix~\ref{app:probe_architectures}. We adopt the standard convention that a multiply-add operation counts as $2$ FLOPs. Throughout this section, we denote the number of denoising steps as $S$, the hidden dimension of the diffusion LLM as $D$, and the average length of the hesitation window for our cascade as $S_{\text{win}}$. The MLP hidden width $K$, attention dimension $d_a$, LSTM projection dimension $d_p$, and LSTM hidden size $d_h$ follow the values specified in Appendix~\ref{app:probe_architectures}. For brevity, we omit lower-order terms such as activation functions (ReLU, tanh, GELU), softmax, layer normalization, and bias additions, which contribute at most $O(D)$ or $O(S_{\text{win}} D)$ FLOPs and are negligible relative to the dominant matrix-vector products.

\paragraph{Linear Probe (LP)}
A single linear probe forward maps a $D$-dimensional vector to a scalar logit, requiring $2D$ FLOPs. The three pooling variants differ in how the trajectory $\mathbf{H} \in \mathbb{R}^{D \times S}$ is reduced to this input. \textbf{LP (Last Step)} uses only $\mathbf{h}_1$, with total cost $2D$. \textbf{LP (Mean)} sums $S$ vectors of dimension $D$ ($SD$ FLOPs) and then applies one LP forward, with total cost $SD + 2D$. \textbf{LP (MV)} applies the LP independently at every denoising step and aggregates by majority vote, with total cost $S \cdot 2D = 2SD$.

\paragraph{MLP}
A single MLP forward consists of $\mathbf{W}_{\text{in}} \in \mathbb{R}^{K \times D}$ and $\mathbf{W}_{\text{out}} \in \mathbb{R}^{1 \times K}$, costing $2DK + 2K \approx 2DK$ FLOPs. \textbf{MLP (Last Step)} applies one MLP forward on $\mathbf{h}_S$, with total cost $2DK$. \textbf{MLP (Mean)} performs mean pooling followed by one MLP forward, with total cost $SD + 2DK$. \textbf{MLP (MV)} applies an MLP forward at every denoising step, with total cost $S \cdot 2DK = 2SDK$.

\paragraph{TimeAttn}
TimeAttn operates on the full trajectory and consists of two main components: per-step attention scoring and a classifier head over the attended representation. The attention scoring computes $\mathbf{W}_a \mathbf{h}_s$ for each step ($\mathbf{W}_a \in \mathbb{R}^{d_a \times D}$), costing $S \cdot 2 D d_a$ FLOPs in total. The weighted sum $\sum_s \alpha_s \mathbf{h}_s$ contributes a further $2SD$ FLOPs. The classifier head is a 2-layer MLP applied to the resulting $D$-dimensional context vector, costing $2DK$ FLOPs. The total cost is therefore $2 S D d_a + 2SD + 2DK \approx 2 S D d_a + 2DK$, where the first term dominates for typical $S$ and $d_a$.

\paragraph{LSTM}
The LSTM probe first projects each $\mathbf{h}_s$ from $D$ to $d_p$ via $\mathbf{W}_{\text{proj}} \in \mathbb{R}^{d_p \times D}$, costing $S \cdot 2 D d_p$ FLOPs across all steps. The projected sequence is then processed by a 2-layer LSTM. Each LSTM cell contains four gates, each with a matrix-vector product of size $d_h \times (d_{\text{in}} + d_h)$, where $d_{\text{in}} = d_p$ for layer 1 and $d_{\text{in}} = d_h$ for layer 2. The total per-step LSTM cost is approximately $8 d_h (d_p + d_h) + 8 d_h \cdot 2 d_h$, applied at all $S$ steps. The final linear head contributes $2 d_h$ FLOPs and is negligible. The total cost is dominated by the input projection, $S \cdot 2 D d_p$, since $D \gg d_h$.

\paragraph{$\mathcal{D}^2$-Monitor (Cascade)}
Our cascade has two components. The base linear probe is applied at every denoising step to compute the per-step margin $|d_s|$ used for hesitation detection, regardless of whether the sample is escalated; this incurs $2SD$ FLOPs per sample. The expert (MLP or TimeAttn) is invoked only on the fraction $p_{\text{esc}}$ of samples flagged as hesitant, and operates on the minimal window covering all hesitation steps with average length $S_{\text{win}} \leq S$, rather than the full trajectory. The expected per-sample cost is
\begin{equation}
    \mathbb{E}[\text{FLOPs}] = \underbrace{2SD}_{\text{base, always}} + \underbrace{p_{\text{esc}} \cdot F_{\text{expert}}(S_{\text{win}})}_{\text{expert, conditional}},
\end{equation}
where the expert FLOPs are
\begin{align}
    F_{\text{MLP-expert}}(S_{\text{win}}) &= S_{\text{win}} D + 2DK \\
    F_{\text{TimeAttn-expert}}(S_{\text{win}}) &= 2 S_{\text{win}} D d_a + 2 S_{\text{win}} D + 2DK.
\end{align}
The values of $p_{\text{esc}}$ and $S_{\text{win}}$ are measured empirically on \textit{WildGuardMix} for each model and reported alongside the FLOPs in Table~\ref{tab:flops}.

Across all four LLaDA models, D$^2$-Monitor achieves a substantially better
efficiency--effectiveness trade-off than full-trajectory baselines.
D$^2$-MLP requires under 1 MFLOP per sample, which is
1.5$\times$--3.3$\times$ cheaper than MLP (Mean), 37$\times$--78$\times$
cheaper than TimeAttn and over 150$\times$ cheaper than LSTM, while
remaining competitive in F1 (\cref{tab:main_results}). D$^2$-TimeAttn is more
expensive than D$^2$-MLP due to the heavier expert, but still costs
3.5$\times$--7$\times$ less than running TimeAttn on the full trajectory.
The savings come from two sources. First, the cascade only invokes the
expert on the fraction $p_\mathrm{esc}$ of samples flagged as hesitant, so
most samples incur only the lightweight $2SD$ FLOPs of the base linear
probe. Second, even on escalated samples, the expert operates on a
localized hesitation window of average length $S_\mathrm{win} \le S$
rather than the full trajectory. As a result, D$^2$-MLP costs less than
even a single-step MLP probe (MLP (Last Step), $\sim$2 MFLOPs) while
attaining performance close to full-trajectory models, demonstrating that
conditional computation effectively decouples cost from accuracy.

\subsection{Disentangling Routing from Hesitation-Curated Training}
\label{app:curation_vs_routing}

$\mathcal{D}^2$-Monitor changes two things at once relative to a standalone
advanced probe: the probe is \emph{trained} on hesitation windows rather than on
full trajectories, and it is \emph{applied} only to the samples the router
escalates. We separate the two factors with a $2\times2$ control over training
data and application on LLaDA-8B-Instruct~(\cref{tab:ablation_2x2}). Two details
make the comparison well posed. First, C2 is evaluated at the same $\lambda$ that
C4 selects, so C2 and C4 escalate an identical set of samples and carry an
identical $\mathrm{E}[P]$; C4 versus C2 therefore isolates the training data at
matched test-time cost. Second, in C3 the hesitation-trained probe is applied to
every test sample: hesitant samples are scored on their hesitation window, while
samples with no hesitation step fall back to the full trajectory.

Routing contributes $+0.64/+0.51$ F1 (MLP/TimeAttn) on WildGuardMix and
$+1.25/+1.58$ on ToxicChat with an all-data probe, and $+0.26/+0.54$ and
$+2.65/+2.83$ respectively with a curated probe. Curated training contributes
$+1.46/+0.99$ and $+0.60/+0.02$ when the probe is applied to every sample, and
$+1.08/+1.02$ and $+2.00/+1.27$ at matched cost. All sixteen contrasts are
non-negative, but the relative size of the two factors reverses between the two
datasets, as does the sign of the interaction term ($-0.38/+0.03$ on
WildGuardMix, $+1.40/+1.25$ on ToxicChat). We therefore report the decomposition
as dataset-dependent rather than claiming a single ordering. What is stable
across both datasets is curated training at matched cost, which is at least
$+1.0$ F1 in all four cases, and routing, which never lowers F1 while reducing
$\mathrm{E}[P]$ by $4.0$--$7.2\times$. These deltas are differences of 3-seed
means and carry no error bars.

\begin{table}[t]
    \centering
    \small
    \setlength{\tabcolsep}{4pt}
    \caption{$2\times2$ ablation over training data (all examples vs.\
    hesitation-curated) and application (all samples vs.\ routed samples), on
    LLaDA-8B-Instruct, mean $\pm$ std over 3 seeds. C2 is evaluated at the same
    $\lambda$ that C4 selects, so C2 and C4 route an identical sample set and
    carry an identical $\mathrm{E}[P]$: C4 vs.\ C2 isolates the training data at
    matched test-time cost. C4 is $\mathcal{D}^2$-Monitor.}
    \label{tab:ablation_2x2}
    \begin{tabular}{lllcccc}
        \toprule
        & \multicolumn{2}{c}{Advanced probe} & \multicolumn{2}{c}{F1 (\%)} & \multicolumn{2}{c}{$\mathrm{E}[P]$} \\
        \cmidrule(lr){2-3} \cmidrule(lr){4-5} \cmidrule(lr){6-7}
        & Training data & Applied to & MLP & TimeAttn & MLP & TimeAttn \\
        \midrule
        \multicolumn{7}{l}{\textit{WildGuardMix}} \\
        C1 & all data & all samples & 87.82$\pm$0.42 & 87.81$\pm$0.21 & 1.05M & 1.59M \\
        C2 & all data & routed      & 88.46$\pm$0.46 & 88.32$\pm$0.30 & 0.26M & 0.22M \\
        C3 & curated  & all samples & 89.28$\pm$0.13 & 88.80$\pm$0.30 & 1.05M & 1.59M \\
        C4 & curated  & routed      & \textbf{89.54$\pm$0.10} & \textbf{89.34$\pm$0.14} & 0.26M & 0.22M \\
        \addlinespace[3pt]
        \multicolumn{7}{l}{\textit{ToxicChat}} \\
        C1 & all data & all samples & 87.08$\pm$0.18 & 87.28$\pm$0.92 & 1.05M & 1.59M \\
        C2 & all data & routed      & 88.33$\pm$0.10 & 88.86$\pm$0.29 & 0.16M & 0.37M \\
        C3 & curated  & all samples & 87.68$\pm$0.93 & 87.30$\pm$1.05 & 1.05M & 1.59M \\
        C4 & curated  & routed      & \textbf{90.33$\pm$0.17} & \textbf{90.13$\pm$0.20} & 0.16M & 0.37M \\
        \bottomrule
    \end{tabular}
\end{table}

\subsection{Constructing the Hesitation Window}
\label{app:window_ablation}
We compare different ways of building the hesitation window~ (\cref{tab:window_ablation}). The default takes the
contiguous span from the first to the last hesitant step, which also contains
the non-hesitant steps in between; \emph{hesitant steps only} drops those,
keeping the same steps but not the contiguity. \emph{Top-$k$} ignores $\tau$
and keeps the $k$ steps with the smallest $|d_t|$, a fixed budget rather than
a data-dependent one ($k\in\{4,8,16,24\}$). \emph{Soft} replaces hard
selection with a weight $\exp(-|d_t|/c\tau)$ on every step, so $c$ controls
how sharply the probe concentrates on low-margin steps
($c\in\{0.25,0.5,1,2\}$); as $c\to\infty$ this becomes a uniform average over
the whole trajectory, which serves as the no-window control. 

What the span buys is consistency, being the best or tied-best variant in
every column, which is why we keep it as the default. How much it buys depends
on the second-stage architecture. $\mathcal{D}^2$-MLP mean-pools its input, so
extra steps are averaged away and most variants land within noise of the span,
a uniform average over the entire trajectory included (90.13 against 90.51).
$\mathcal{D}^2$-TimeAttn attends over the sequence and registers what is in
it: the uniform average costs 0.69 points, and taking the same number of steps
non-contiguously by margin (top-$k$, $k=16$) costs 0.67. 

\subsection{Comparison with LLM-based Monitors}
\label{app:llm_monitors}

\begin{table}[t]
    \centering
    \small
    \setlength{\tabcolsep}{5pt}
    \begin{tabular}{lccccccc}
        \toprule
        & & \multicolumn{3}{c}{$\mathcal{D}^2$-MLP} & \multicolumn{3}{c}{$\mathcal{D}^2$-TimeAttn} \\
        \cmidrule(lr){3-5}\cmidrule(lr){6-8}
        Window construction & $|w|$ & $\lambda^\star$ & F1 @ $\lambda^\star$ & F1 @ $\lambda_{\mathrm{ref}}$
                                   & $\lambda^\star$ & F1 @ $\lambda^\star$ & F1 @ $\lambda_{\mathrm{ref}}$ \\
        \midrule
        \multicolumn{8}{l}{\textit{Threshold-based}} \\
        \rowcolor{aliceblue}
        \quad Hesitation span (default) & 16.9 & 24 & \textbf{90.51} & \textbf{90.51}
                                               & 4  & \textbf{90.35} & \textbf{90.35} \\
        \quad Hesitant steps only       & 12.9 & 29 & 90.05 & 90.00 & 4  & 90.22 & 90.22 \\
        \addlinespace[2pt]
        \multicolumn{8}{l}{\textit{Fixed budget}} \\
        \quad Top-$k$, $k=4$            &  4.0 & 29 & 89.74 & 89.76 & 29 & 89.72 & 89.14 \\
        \quad Top-$k$, $k=8$            &  8.0 & 29 & 89.91 & 89.76 & 21 & 89.81 & 89.66 \\
        \quad Top-$k$, $k=16$           & 16.0 & 22 & 90.28 & 90.16 & 21 & 89.75 & 89.68 \\
        \quad Top-$k$, $k=24$           & 24.0 & 17 & 90.42 & 90.20 & 17 & 89.99 & 89.71 \\
        \addlinespace[2pt]
        \multicolumn{8}{l}{\textit{Soft weighting}} \\
        \quad Soft, $c=0.25$            & 13.0 & 29 & 90.05 & 89.88 & 21 & 89.85 & 89.91 \\
        \quad Soft, $c=0.5$             & 21.8 & 29 & 90.17 & 90.12 & 19 & 89.41 & 89.01 \\
        \quad Soft, $c=1$               & 28.2 & 13 & 90.08 & 89.86 & 18 & 89.74 & 89.46 \\
        \quad Soft, $c=2$               & 30.9 & 21 & 90.20 & 90.13 & 18 & 90.21 & 90.21 \\
        \midrule
        Full trajectory (uniform, $c\!\to\!\infty$) & 32.0 & 15 & 90.32 & 90.13 & 17 & 90.22 & 89.66 \\
        \bottomrule
    \end{tabular}
    \caption{\textbf{Ablation over hesitation-window construction} on ToxicChat
    with LLaDA-8B-Instruct (F1 in \%, seed 0). Only the window rule changes;
    everything else is held fixed. $|w|$ is the mean effective window length in
    steps (out of $S=32$), identical for both probes since both the window rule
    and the margins are shared. We report each variant under two routing
    thresholds. $\lambda^\star$ is the threshold selected for that variant by
    the same rule as the main results, and therefore differs across rows.
    $\lambda_{\mathrm{ref}}$ is the default window's $\lambda^\star$ ($24$ for
    $\mathcal{D}^2$-MLP, $4$ for $\mathcal{D}^2$-TimeAttn), held fixed across
    rows. Because routing depends on the number of hesitant steps rather than
    on the window rule, every row at $\lambda_{\mathrm{ref}}$ escalates the
    same samples at the same $\mathrm{E}[P]$ ($140.5$K and $629.7$K
    respectively), which makes that column a cost-matched comparison. Best per
    column in \textbf{bold}.}
    \label{tab:window_ablation}
\end{table}

\paragraph{Setup}
We finetune Llama-Guard-3-1B
\citep{metallamaguard3} is against its own prompt template, which we
apply unmodified through \texttt{tokenizer.apply\_chat\_template}. The template
asks the model to check the conversation against the MLCommons hazard taxonomy
(S1--S13: violent crimes, non-violent crimes, sex-related crimes, child sexual
exploitation, defamation, specialized advice, privacy, intellectual property,
indiscriminate weapons, hate, suicide and self-harm, sexual content, and
elections) and to answer \texttt{safe} or \texttt{unsafe} on the first line,
followed on a second line by the violated category keys. We keep the default
category set and treat \texttt{unsafe} as a positive detection.

We evaluate two configurations. \emph{Zero-shot} applies the released
checkpoint directly. \emph{Aligned} adapts it with LoRA on the training split
of each dataset, using the same splits as the probe-based monitors so that both
sides see identical training data: rank $16$, $\alpha=32$, dropout $0.05$,
applied to all attention and MLP projection matrices, learning rate
$2\times10^{-4}$ with a linear schedule, and effective batch size $16$.

\paragraph{What the two monitors have access to}
The comparison is not symmetric, and the asymmetry cuts both ways. Llama-Guard
classifies the input text directly, so it needs no access to the model being
monitored and a single row applies to all four D-LLMs, whereas
$\mathcal{D}^2$-Monitor reads internal states and is trained per model. In
exchange, $\mathcal{D}^2$-Monitor rides on a forward pass that is already
happening, while Llama-Guard adds an independent model call to every request.
Which side this favors depends on the deployment: a provider serving many
different models may prefer one text-level monitor, while a setting with a
fixed model and a tight per-request budget favors the probe.

\paragraph{Results}
\cref{tab:llamaguard} reports the full metric breakdown and
\cref{tab:llm_monitor_comparison} places the two monitors side by side at
matched training data. Three points stand out. First, without task-specific
adaptation Llama-Guard is substantially weaker, reaching 60.7 F1 on ToxicChat
at 21.8 precision. Second, once aligned it matches $\mathcal{D}^2$-Monitor
in-domain, but does so with over $2000\times$ as many monitor parameters at
inference. Third, under transfer it retains more of its performance than any
probe-based method in \cref{tab:cross_dataset}, though not uniformly: on
OpenAI-Moderation, alignment adds only 0.9 F1 over zero-shot while raising the
false-flag rate on safe inputs from 26.8\% to 32.7\%.



\begin{table}[t]
\centering
\small
\setlength{\tabcolsep}{6pt}
\caption{Zero-shot vs.\ LoRA-aligned Llama-Guard-3-1B on binary safe/unsafe
prompt classification (\%). Unsafe is the positive class; FRR is the false-flag
rate on truly safe inputs. $^\dagger$ marks in-domain evaluation (eval split
matches the LoRA's training set); all other LoRA rows are out-of-distribution.
Note that on OpenAI-Moderation, alignment leaves accuracy unchanged and raises
FRR.}
\label{tab:llamaguard3-1b-align}
\begin{tabular}{llccccc}
\toprule
Eval data & Model & Acc. & F1 & Prec. & Rec. & FRR $\downarrow$ \\
\midrule
\multirow{2}{*}{WildGuardMix}
  & Zero-shot                     & 79.5 & 78.8 & 82.4 & 68.4 & 11.6 \\
  & LoRA (WildGuardMix)$^\dagger$ & 90.1 & 89.9 & 92.1 & 85.0 & \phantom{0}5.8 \\
\addlinespace[4pt]
\multirow{2}{*}{OpenAI-Mod}
  & Zero-shot                     & 75.3 & 73.6 & 57.4 & 79.9 & 26.8 \\
  & LoRA (WildGuardMix)           & 75.2 & 74.5 & 56.1 & 92.9 & 32.7 \\
\addlinespace[4pt]
\multirow{3}{*}{ToxicChat}
  & Zero-shot                     & 82.7 & 60.7 & 21.8 & 55.5 & 15.2 \\
  & LoRA (WildGuardMix)           & 93.2 & 80.4 & 51.4 & 86.7 & \phantom{0}6.3 \\
  & LoRA (ToxicChat)$^\dagger$    & 97.5 & 90.4 & 83.2 & 80.9 & \phantom{0}1.2 \\
\bottomrule
\label{tab:llamaguard}
\end{tabular}
\end{table}

\begin{table}[t]
    \centering
    \small
    \setlength{\tabcolsep}{5pt}
    \begin{tabular}{lcccc}
        \toprule
        & \multicolumn{2}{c}{Llama-Guard-3-1B} & \multicolumn{2}{c}{$\mathcal{D}^2$-Monitor} \\
        \cmidrule(lr){2-3}\cmidrule(lr){4-5}
        & zero-shot & LoRA & $\mathcal{D}^2$-MLP & $\mathcal{D}^2$-TimeAttn \\
        \midrule
        \multicolumn{5}{l}{\textit{Adapted on WildGuardMix}} \\
        \quad Monitor params            & 1.12B & 1.12B & $\leq$0.34M & $\leq$0.55M \\
        \quad WildGuardMix$^\dagger$    & 78.8  & 89.9  & 89.5 & 89.3 \\
        \quad ToxicChat                 & 60.7  & 80.4  & 73.2 & 75.3 \\
        \quad OpenAI-Mod                & 73.6  & 74.5  & 68.9 & 70.5 \\
        \addlinespace[3pt]
        \multicolumn{5}{l}{\textit{Adapted on ToxicChat}} \\
        \quad Monitor params            & 1.12B & 1.12B & $\leq$0.18M & $\leq$0.37M \\
        \quad ToxicChat$^\dagger$       & 60.7  & 90.4  & 90.3 & 90.1 \\
        \bottomrule
    \end{tabular}
    \caption{$\mathcal{D}^2$-Monitor versus an LLM-based monitor at matched
    adaptation data. Entries are F1 (\%) except the parameter rows.
    $^\dagger$ marks in-domain evaluation; all other entries are transfer. The
    zero-shot column is the released Llama-Guard checkpoint and does not depend
    on the adaptation set, so it is repeated in both blocks as a reference
    point. $\mathcal{D}^2$-Monitor rows are LLaDA-8B-Instruct; the parameter count is $\mathrm{E}[P]$ for the in-domain setting (marked
$^\dagger$); values for the transfer settings are in
\cref{tab:cross_dataset}. We did
    not evaluate transfer from ToxicChat, so that block reports in-domain
    results only.}
    \label{tab:llm_monitor_comparison}
\end{table}

\section{Additional Analysis}
\subsection{Analysis of Hesitation Dynamics}
\label{app:hesitation_analysis}

\begin{figure*}[p]
\centering
\newcommand{\subw}{0.485\textwidth}
\newcommand{\subh}{0.18\textheight}

\begin{subfigure}[t]{\subw}
    \centering
    \includegraphics[width=\linewidth,height=\subh,keepaspectratio]{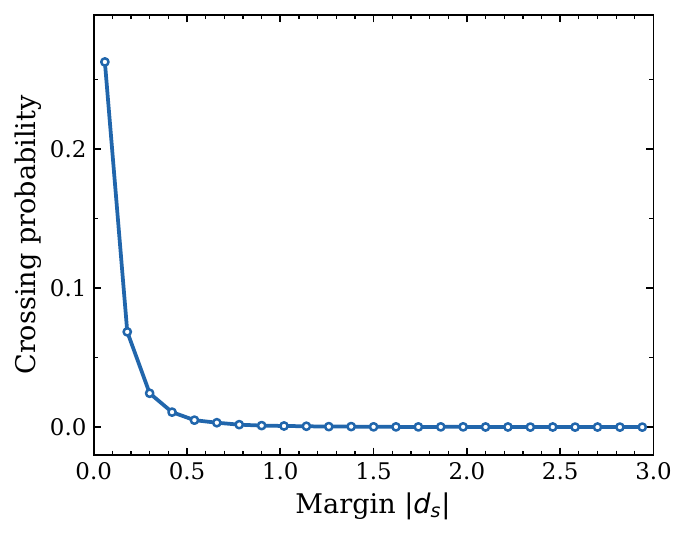}
    \caption{LLaDA-8B-Base: Crossing}
\end{subfigure}
\hfill
\begin{subfigure}[t]{\subw}
    \centering
    \includegraphics[width=\linewidth,height=\subh,keepaspectratio]{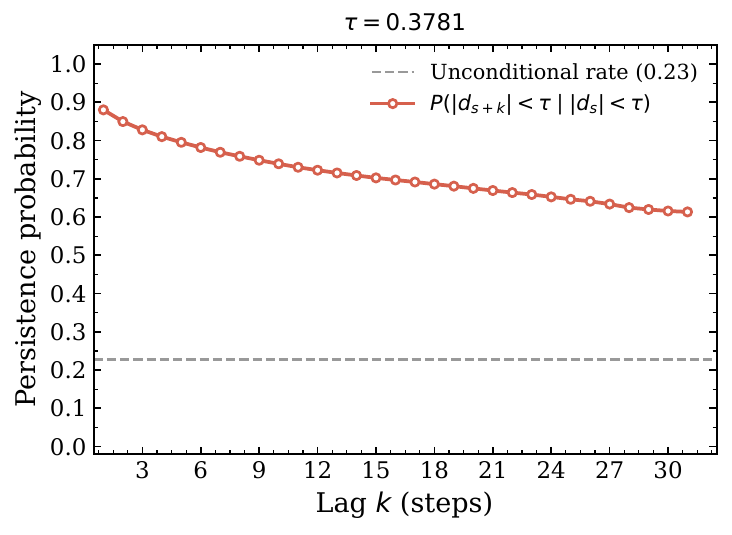}
    \caption{LLaDA-8B-Base: Persistence}
\end{subfigure}

\vspace{4pt}

\begin{subfigure}[t]{\subw}
    \centering
    \includegraphics[width=\linewidth,height=\subh,keepaspectratio]{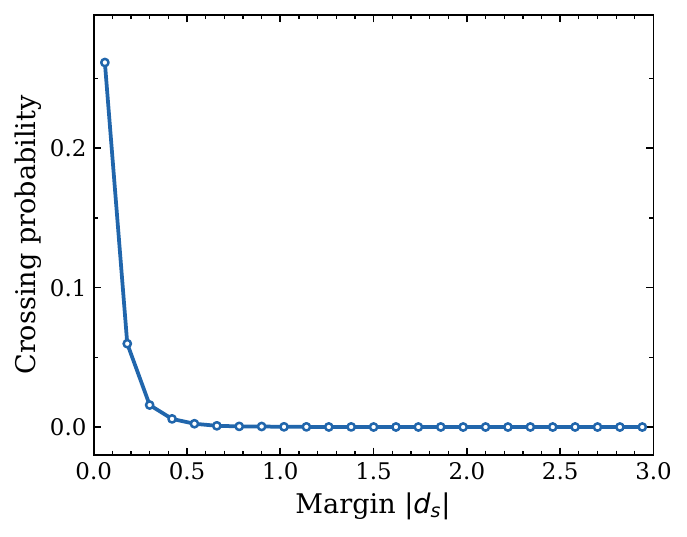}
    \caption{LLaDA-8B-Instruct: Crossing}
\end{subfigure}
\hfill
\begin{subfigure}[t]{\subw}
    \centering
    \includegraphics[width=\linewidth,height=\subh,keepaspectratio]{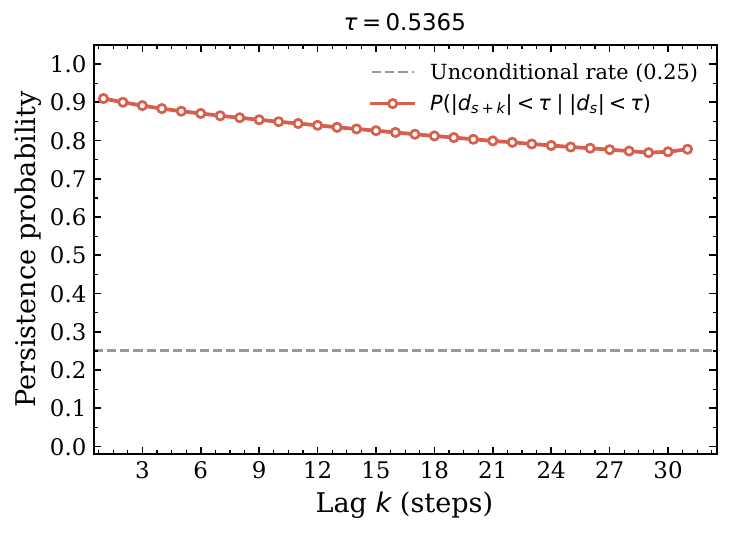}
    \caption{LLaDA-8B-Instruct: Persistence}
\end{subfigure}

\vspace{4pt}

\begin{subfigure}[t]{\subw}
    \centering
    \includegraphics[width=\linewidth,height=\subh,keepaspectratio]{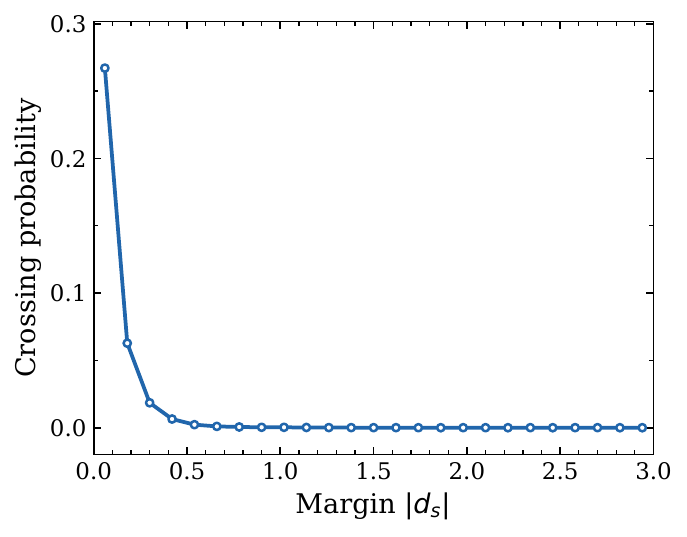}
    \caption{LLaDA-1.5: Crossing}
\end{subfigure}
\hfill
\begin{subfigure}[t]{\subw}
    \centering
    \includegraphics[width=\linewidth,height=\subh,keepaspectratio]{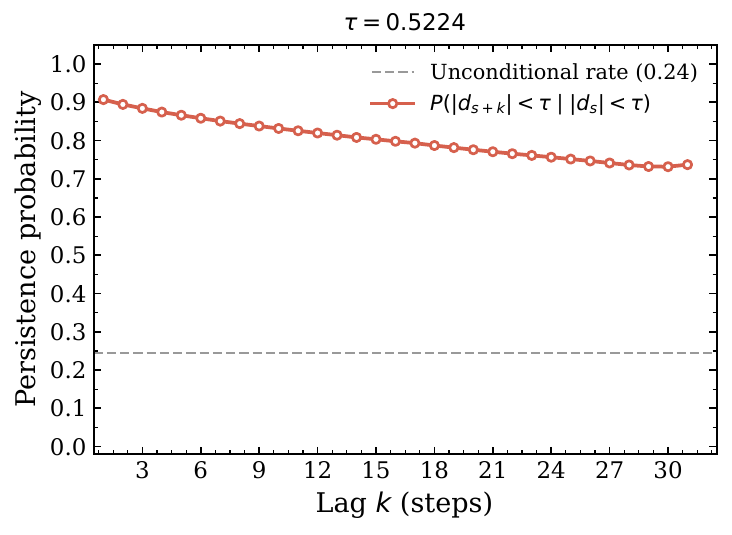}
    \caption{LLaDA-1.5: Persistence}
\end{subfigure}

\vspace{4pt}

\begin{subfigure}[t]{\subw}
    \centering
    \includegraphics[width=\linewidth,height=\subh,keepaspectratio]{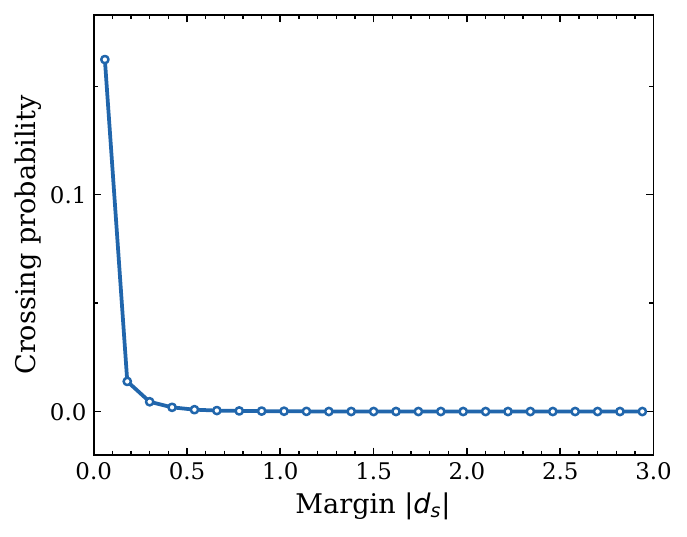}
    \caption{LLaDA-2.0-mini: Crossing}
\end{subfigure}
\hfill
\begin{subfigure}[t]{\subw}
    \centering
    \includegraphics[width=\linewidth,height=\subh,keepaspectratio]{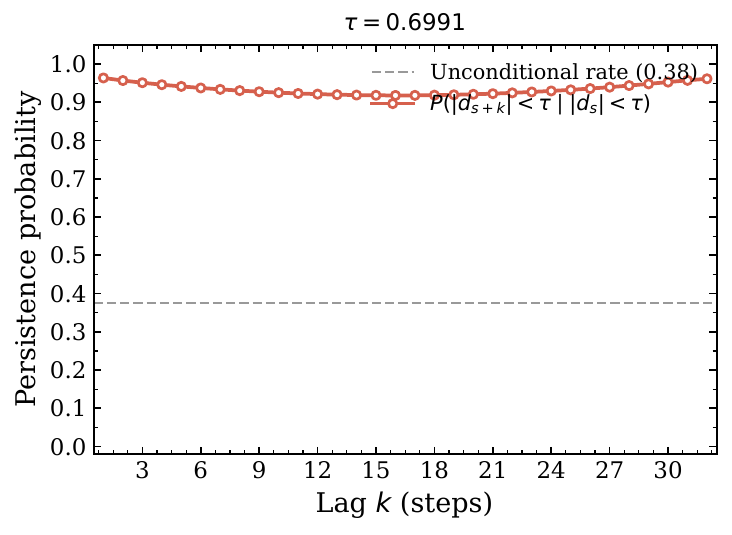}
    \caption{LLaDA-2.0-mini: Persistence}
\end{subfigure}

\caption{
Cross-boundary probability (left) and margin persistence (right) across four LLaDA variants. Left: crossing probability as a function of margin $|d_s|$. Right: persistence probability $P(|d_{s+k}| < \tau \mid |d_s| < \tau)$. Margins are determined by OOF scoring on \textit{WildGuardMix}.
}
\label{fig:hesitation_analysis}
\end{figure*}

\paragraph{Cross-Boundary Probability}
We first analyze the local instability of model predictions by measuring how likely a step is to cross the decision boundary in the next step. 
Given the signed margin $d_s$ at step $s$, we compute the probability that the sign of the margin changes at the next step, i.e., $\mathrm{sign}(d_{s+1}) \neq \mathrm{sign}(d_s)$, 
as a function of the margin $|d_s|$.
Figure~\ref{fig:hesitation_analysis} shows the crossing probability binned over $|d_s|$. 
Across all four LLaDA models, we observe a consistent pattern: 
steps close to the decision boundary exhibit a significantly higher probability of crossing it in the next step, 
while steps far from the boundary are highly stable. 
The crossing probability decreases sharply as $|d_s|$ increases, quickly approaching zero beyond a small margin threshold. 
This indicates that hesitation steps, characterized by small margin magnitudes, are intrinsically unstable and prone to prediction flips, 
suggesting that the decision boundary region concentrates most of the local uncertainty in the trajectory.

\paragraph{Margin Persistence}
\label{Margin Persistence}
We further analyze the temporal structure of hesitation by measuring how long a model remains in a hesitant state. 
Given a threshold $\tau$, we define a step as hesitant if $|d_s| < \tau$. 
For each such step, we compute the probability that the model remains hesitant after $k$ steps, i.e.,
\[
P(|d_{s+k}| < \tau \mid |d_s| < \tau),
\]
for $k = 1, 2, \ldots, K$. \cref{fig:hesitation_analysis} shows the persistence curves for different models. 
We also report the unconditional probability of being in a hesitant state as a baseline. 
Across models, we observe that hesitation exhibits clear temporal persistence: 
once the model enters a low-margin region, it tends to remain in that region for multiple subsequent steps, 
with the persistence probability decaying gradually as $k$ increases.
These results suggest that hesitation is not only a local phenomenon but also forms contiguous segments along the generation trajectory. 
This temporal coherence further motivates our design of operating on localized hesitation windows, 
as they capture the regions where uncertainty is both concentrated and temporally structured.

\subsection{ Margin Outperforms Entropy and Confidence as Step-Count Signal}
\label{app:step_count_signals}
\begin{figure}[t]
\centering
\begin{subfigure}[b]{0.48\linewidth}
    \includegraphics[width=\linewidth]{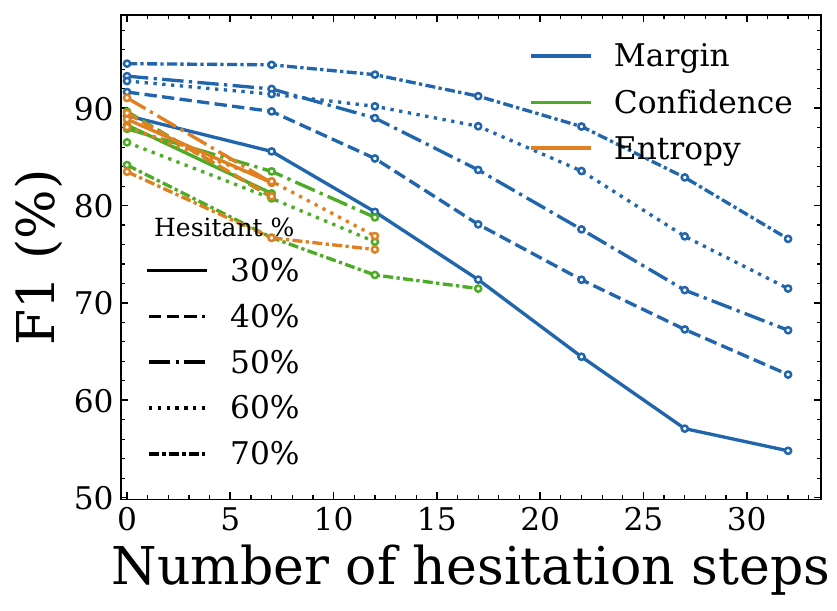}
    \caption{LP (MV)}
    \label{fig:step_count_compare:mv}
\end{subfigure}
\hfill
\begin{subfigure}[b]{0.48\linewidth}
    \includegraphics[width=\linewidth]{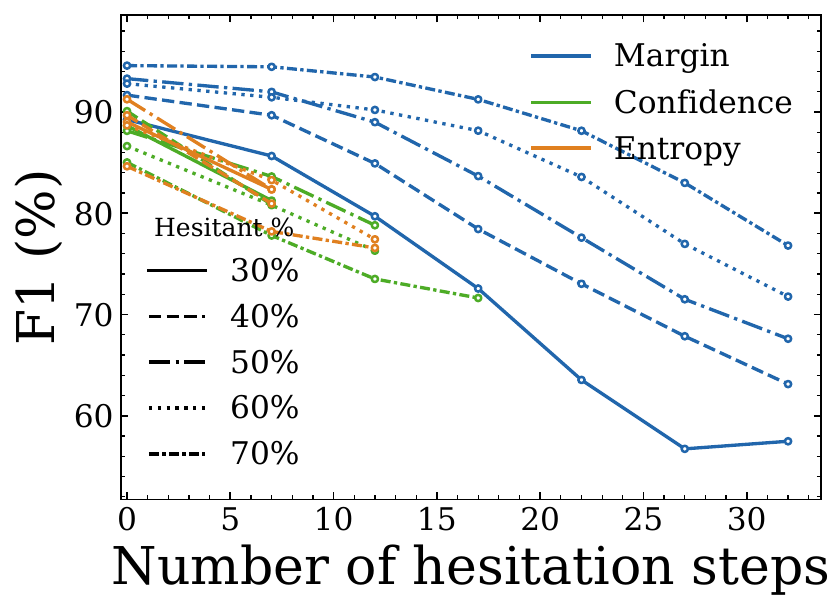}
    \caption{LP (Mean)}
    \label{fig:step_count_compare:mean}
\end{subfigure}
\caption{\textbf{Comparison of step-count signals on LLaDA-8B-Instruct.} For each step-count signal ($n_\tau$ from probe margin, $n_{\text{entropy}}$ from step-wise entropy, $n_{\text{confidence}}$ from step-wise confidence), we report probe F1 across buckets of increasing hesitation count, under two base classifier variants: \textbf{(a)} LP (MV) and \textbf{(b)} LP (Mean). Each color corresponds to one signal, and each line style corresponds to a different threshold setting parameterized by the resulting hesitant ratio. Across both base classifiers, the margin-based signal $n_\tau$ produces the steepest and most extended F1 decline, indicating it stratifies difficulty more discriminatively than the probe-extrinsic signals.}
\label{fig:step_count_compare}
\end{figure}

In Section~\ref{sec:trajectory-as-signal}, we noted that the step-count construction underlying $n_\tau$ is general and can be applied to any per-step hesitation signal. We instantiate this for the two probe-extrinsic signals introduced in Section~\ref{hesitaion severity}, defining
\begin{align}
    n_{\text{entropy}} &= \sum_{s=1}^{S} \mathbb{I}[E_s \geq \tau_E], \\
    n_{\text{confidence}} &= \sum_{s=1}^{S} \mathbb{I}[C_s \leq \tau_C],
\end{align}
where $E_s$ and $C_s$ are the step-wise entropy and confidence scores, and $\tau_E, \tau_C$ are their respective thresholds. We compare the three step-count signals on LLaDA-8B-Instruct in~\cref{fig:step_count_compare}, where for each signal we evaluate probe F1 across $n_\tau$ buckets (more generally, $n_{\text{signal}}$ buckets) under five threshold settings parameterized by the resulting hesitant ratio.
Two observations stand out. First, all three signals are predictive of difficulty in the qualitative sense: F1 decreases monotonically as the count increases, regardless of which signal is used. This confirms that the step-count construction itself is a general approach to extracting trajectory-level difficulty information from per-step signals.
Second, $n_\tau$ produces the steepest and most extended F1 decline across all hesitant ratios. The margin-based curves (blue) drop from $\sim 90\%$ at $n_\tau = 0$ to $55$--$77\%$ at the largest buckets, whereas entropy-based and confidence-based curves (orange and green) plateau at $\sim 75$--$80\%$ and cover a much narrower range of $n_{\text{signal}}$ values. The larger F1 spread under $n_\tau$ indicates that the probe margin produces a finer and more discriminative stratification of difficulty than entropy or confidence: trajectories deemed "highly hesitant" under $n_\tau$ are substantially harder for the probe than those deemed "highly hesitant" under entropy or confidence. This justifies our choice of $n_\tau$ as the routing signal in $\mathcal{D}^2$-Monitor.
A natural explanation is that the probe margin is \emph{probe-aware}: it directly measures distance to the probe's decision boundary, which is the quantity governing whether the probe will misclassify a sample. In contrast, entropy and confidence are computed from the D-LLM's predicted token distribution and are agnostic to the specific probe being used. They capture model-level uncertainty but miss the probe-specific decision dynamics that matter for routing.

\subsection{Hesitation Severity Captures Adversarial Inputs}
\label{app:hesitation_adv}
\begin{figure}[t]
\centering
\begin{subfigure}[b]{0.32\linewidth}
    \includegraphics[width=\linewidth]{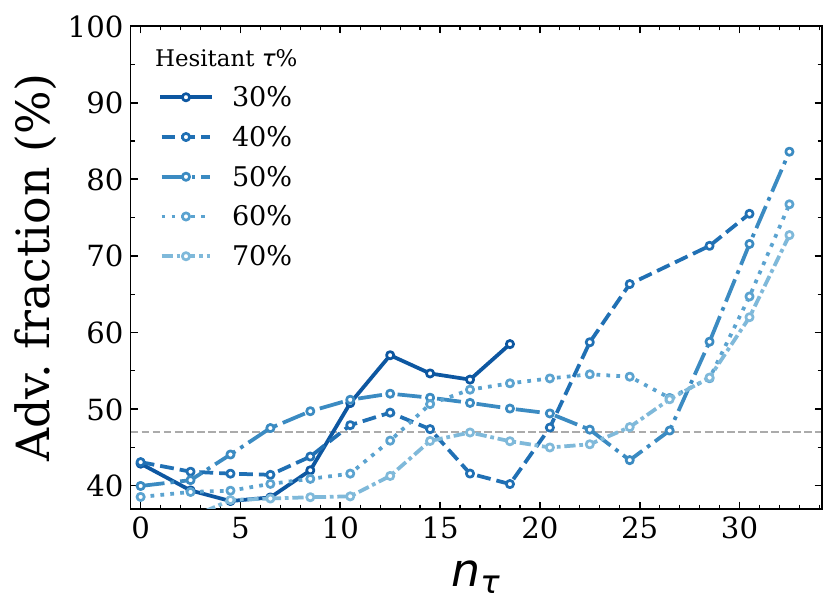}
    \caption{LLaDA-8B-Base}
    \label{fig:hes_adv:base}
\end{subfigure}
\hfill
\begin{subfigure}[b]{0.32\linewidth}
    \includegraphics[width=\linewidth]{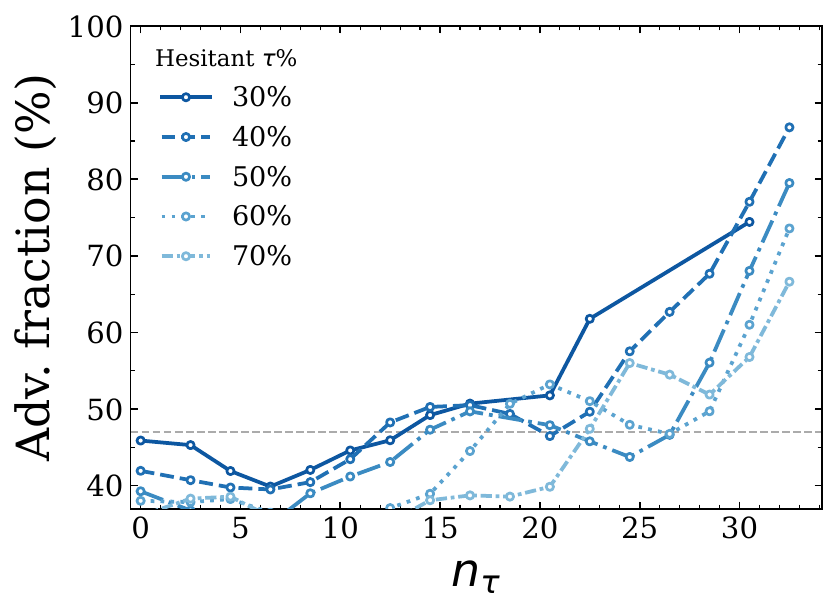}
    \caption{LLaDA-8B-Instruct}
    \label{fig:hes_adv:ins}
\end{subfigure}
\hfill
\begin{subfigure}[b]{0.32\linewidth}
    \includegraphics[width=\linewidth]{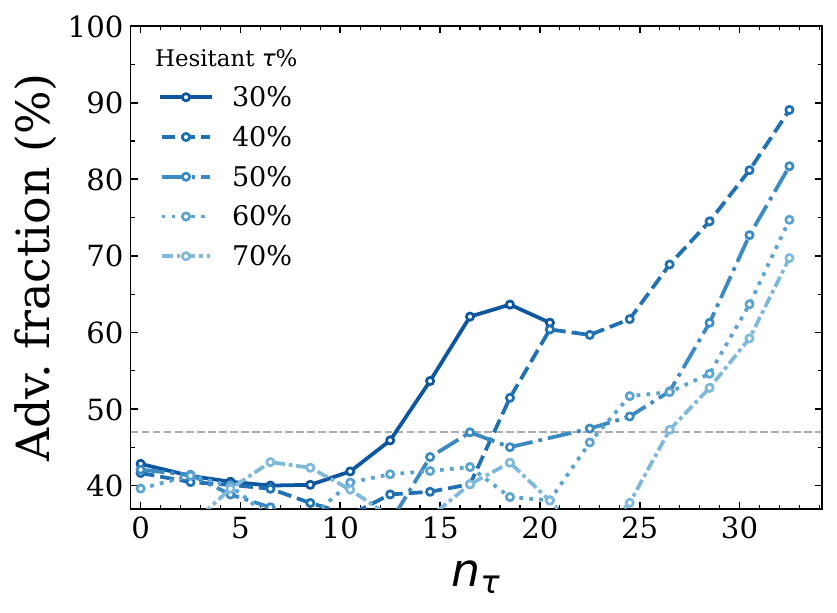}
    \caption{LLaDA-1.5}
    \label{fig:hes_adv:1_5}
\end{subfigure}
\caption{\textbf{Adversarial fraction vs.\ hesitation severity $n_\tau$.} For each $n_\tau$ bucket, we report the fraction of samples drawn from the adversarial split of \textit{WildGuardMix}. Each curve corresponds to a different $\tau$ setting (parameterized by the resulting hesitant ratio). The gray dashed line marks the dataset-wide adversarial fraction ($\sim 47\%$). Adversarial fraction increases with $n_\tau$ on all three LLaDA variants, starting
below the dataset baseline and ending well above it.}
\label{fig:hes_adv}
\end{figure}

We further examine \emph{which} samples accumulate large $n_\tau$, beyond their being difficult for the linear probe. Recall that \textit{WildGuardMix} contains a mixture of natural and adversarially designed prompts, where the latter are constructed to evade safety classifiers and are widely regarded as the harder portion of the benchmark~\cite{han2024wildguard}. We compute, for each $n_\tau$ bucket, the fraction of samples drawn from the adversarial split (\textit{Adv.~fraction}) and report the result on three LLaDA variants in~\cref{fig:hes_adv}. The relationship is consistent across all three models. At $n_\tau = 0$, the adversarial fraction is $38$--$46\%$, noticeably below the dataset-wide baseline of $\sim 47\%$ (gray dashed line); as $n_\tau$ increases, the fraction rises monotonically and reaches $67$--$89\%$ at the largest buckets across the five $\tau$ settings. The overall trend, including both the sub-baseline behavior at $n_\tau = 0$ and the monotonic rise toward $n_\tau \to S$, holds for LLaDA-8B-Base, LLaDA-8B-Instruct, and LLaDA-1.5, indicating that the association between hesitation severity and adversarial inputs is a model-agnostic property rather than an artifact of any specific model or threshold choice. Conceptually, this is consistent with the design intent of adversarial prompts: they are crafted to push the model into a borderline decision, which manifests as repeated proximity to the probe's decision boundary across denoising steps. Hesitation severity $n_\tau$ thus naturally captures this signature of adversarial conditioning, providing a concrete semantic interpretation beyond abstract probe uncertainty.
The above analysis is conducted at the level of $n_\tau$ buckets and characterizes the data property of hesitation. We now turn to the operational consequence of this property in our cascade: \emph{which samples are actually routed to the second-stage probe?} Since routing in $\mathcal{D}^2$-Monitor is triggered by $n_\tau$ exceeding a threshold (selected on a held-out validation set), the routed subset should, by construction, inherit the adv-rich property documented above. We empirically verify this by computing the adversarial fraction within the routed subset for each LLaDA variant and report the result in~\cref{tab:routed_adv}. The routed subset is consistently and substantially enriched in adversarial inputs relative to the dataset-wide baseline of $\sim 47\%$: the adversarial fraction reaches $68.2\%$ on LLaDA-8B-Base, $80.8\%$ on LLaDA-8B-Instruct, $80.5\%$ on LLaDA-1.5, and $75.5\%$ on LLaDA-2.0-mini for $\mathcal{D}^2$-MLP, with $\mathcal{D}^2$-TimeAttn exhibiting comparable enrichment ($64.4\%$, $87.5\%$, $80.5\%$ ,$75.5\%$, respectively). In other words, the cascade routes predominantly adversarial samples to the expert. This confirms that $\mathcal{D}^2$-Monitor does not merely allocate extra capacity to ``hard'' samples in a generic sense but specifically channels it toward adversarial inputs that pose the highest risk to safety classification. Viewed in this light, the bi-level design becomes more than a cost-saving heuristic: it is a targeted defense mechanism, with hesitation severity acting as an implicit detector of adversarial conditioning, and the second-stage expert serving as the specialized classifier for these flagged cases.

\begin{figure}[t]
    \centering
    \includegraphics[width=0.6\linewidth]{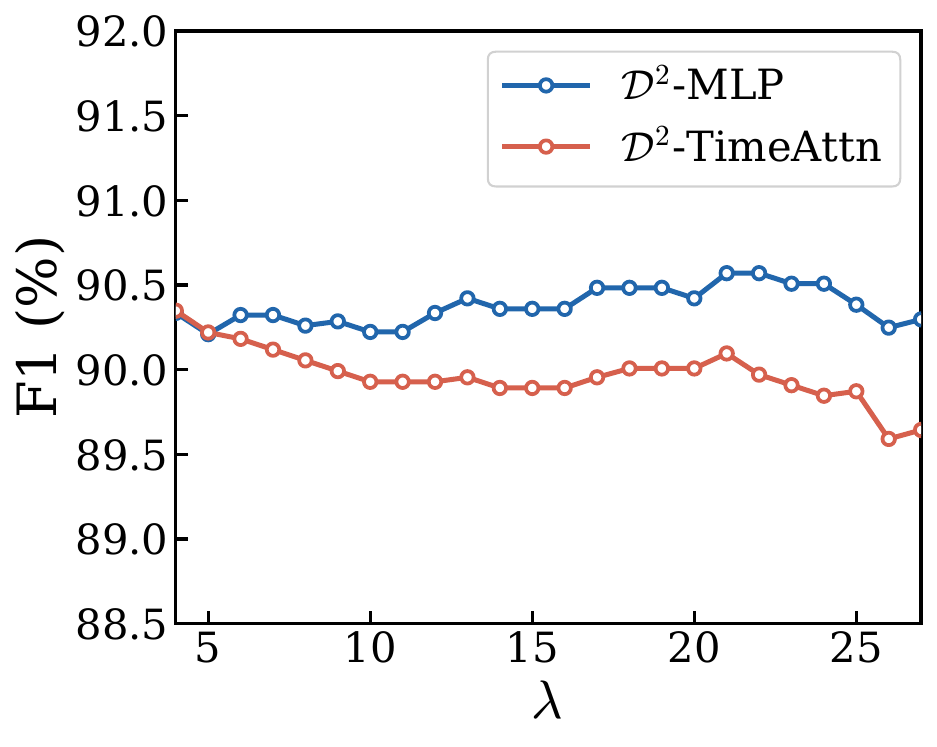}
    \caption{In-distribution sensitivity of F1 to $\lambda$ for
    $\mathcal{D}^2${-Monitor} on ToxicChat (LLaDA-8B-Instruct), with $\tau$
    fixed at its training value. F1 varies by less than 0.8 points across
    $\lambda \in [4, 27]$ for both probe architectures.}
    \label{fig:mdt_sensitivity_indomain}
\end{figure}

\subsection{Sensitivity to Test-time Hyperparameters}
\label{app:sensitivity}

Both $\tau$ and $\lambda$ are selected at test time and require no retraining,
so a practitioner can move along the resulting operating curve at no additional
training cost. We therefore sweep them directly on the trained monitors. For
in-distribution evaluation we keep $\tau$ at the value selected on the training
data and sweep only $\lambda$, since the test distribution matches the training
one. For cross-dataset evaluation, we sweep $\tau$ and $\lambda$ jointly.

\cref{fig:mdt_sensitivity_indomain} reports the in-distribution sweep. F1 varies
by less than 0.8 points across $\lambda \in [4, 27]$ for both probe
architectures, so $\lambda$ can be traded against $\mathrm{E}[P]$ over a wide
range at little cost in detection quality.

\begin{figure*}[t]
    \centering
    \begin{subfigure}[b]{0.48\textwidth}
        \centering
        \includegraphics[width=\linewidth]{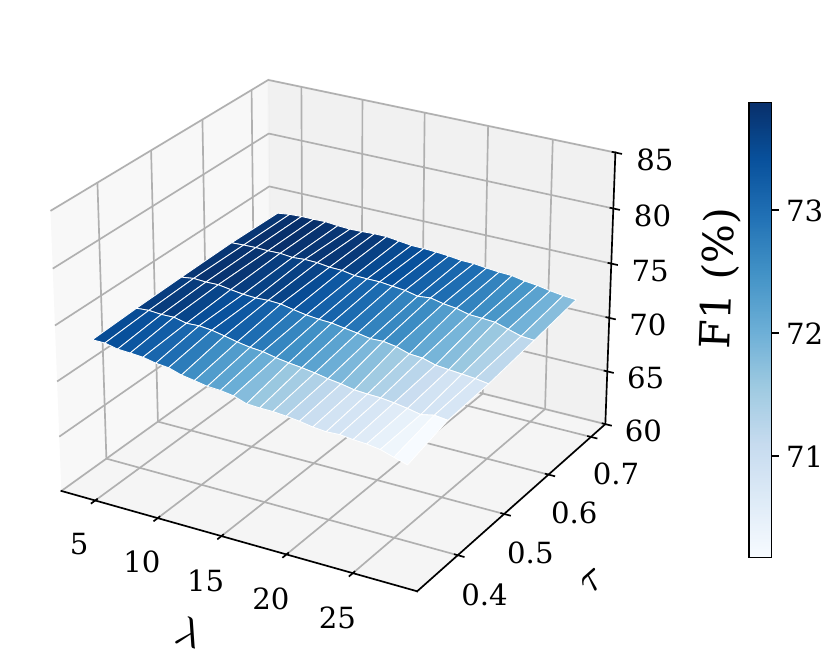}
        \caption{$\mathcal{D}^2$-MLP}
        \label{fig:tau_mdt_cross_mlp}
    \end{subfigure}
    \hfill
    \begin{subfigure}[b]{0.48\textwidth}
        \centering
        \includegraphics[width=\linewidth]{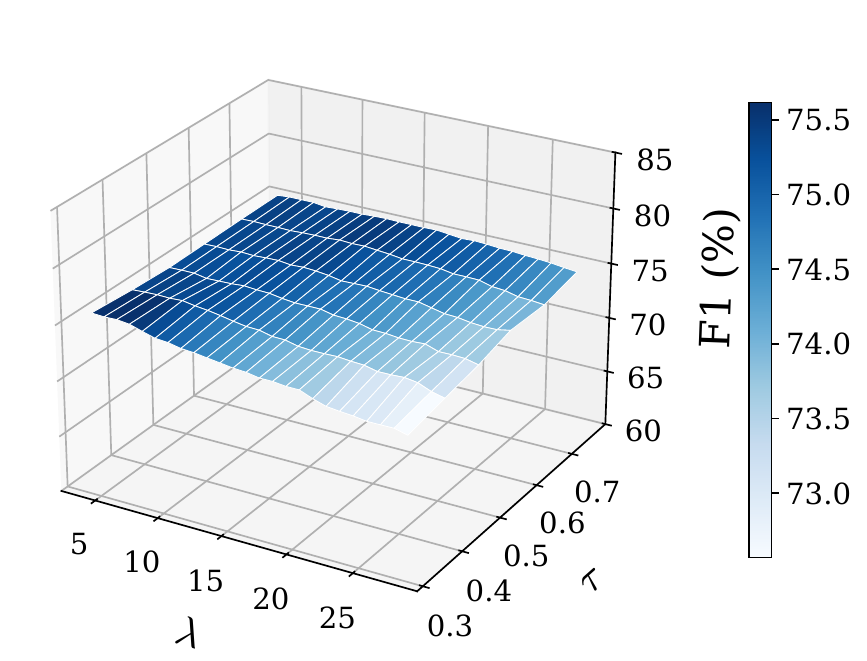}
        \caption{$\mathcal{D}^2$-TimeAttn}
        \label{fig:tau_mdt_cross_timeattn}
    \end{subfigure}
    \caption{Cross-dataset sensitivity of F1 to $(\tau, \lambda)$ for
    $\mathcal{D}^2${-Monitor} trained on WildGuardMix and evaluated on ToxicChat
    (LLaDA-8B-Instruct). Both $\tau$ and $\lambda$ are varied at test time
    without retraining.}
    \label{fig:tau_mdt_sensitivity_cross}
\end{figure*}

\cref{fig:tau_mdt_sensitivity_cross} reports the cross-dataset sweep. Over the
entire $(\tau, \lambda)$ grid, F1 spans 3.5 points for $\mathcal{D}^2$-MLP and
3.3 points for $\mathcal{D}^2$-TimeAttn, indicating that the cross-dataset
results in \cref{tab:cross_dataset} do not depend on a narrow choice of
hyperparameters.

\subsection{Sensitivity to the Probing Layer}
\label{app:layer_sensitivity}

The hidden states we probe are taken from a single layer of the D-LLM. We
check how much this choice matters by repeating the evaluation with probes
trained on layers 20, 24, 28 and 32 of LLaDA-8B-Instruct, keeping the rest of
the pipeline fixed. Unlike the
multi-seed results elsewhere in this appendix, these numbers come from a
single run (seed 0).

Across the four mid-to-late layers, F1 varies by about 0.5 points for both
variants (89.2--89.7 for $\mathcal{D}^2$-MLP and 89.1--89.4 for
$\mathcal{D}^2$-TimeAttn), so the method does not depend on the exact layer.
This is consistent with recent evidence that diffusion LLMs carry substantially
redundant representations across layers, with many layers encoding similar
information \citep{goel2026skip}.

\begin{table}[h]
\centering
\setlength{\tabcolsep}{14pt}
\renewcommand{\arraystretch}{1.2}
\caption{\textbf{Sensitivity to the probing layer.} F1 on
\textit{WildGuardMix} with LLaDA-8B-Instruct, probes taken from different
layers. Single run (seed 0); all other settings follow
Appendix~\ref{app:exp}.}
\label{tab:layer_sensitivity}
\begin{tabular}{ccc}
\toprule
Probing layer & $\mathcal{D}^2$-MLP & $\mathcal{D}^2$-TimeAttn \\
\midrule
20 & 89.2 & 89.1 \\
24 & 89.3 & 89.1 \\
28 & 89.2 & 89.2 \\
32 & 89.7 & 89.4 \\
\bottomrule
\end{tabular}
\end{table}

\clearpage

\begin{table}[t]
\centering
\caption{\textbf{Adversarial fraction in the routed subset.} For each LLaDA
variant, we report the percentage of routed samples that are drawn from the
adversarial split of \textit{WildGuardMix}, against a dataset-wide baseline of
$\sim 47\%$. Routing thresholds are selected on a held-out validation set per
method.}
\label{tab:routed_adv}
\setlength{\tabcolsep}{6pt}
\renewcommand{\arraystretch}{1.15}
\begin{tabular}{lcccc}
\toprule
\textbf{Method} & \textbf{LLaDA-8B-Base} & \textbf{LLaDA-8B-Instruct} &
\textbf{LLaDA-1.5} & \textbf{LLaDA-2.0-mini} \\
\midrule
$\mathcal{D}^2$-MLP      & 68.2\% & 80.8\% & 80.5\% & 75.5\% \\
$\mathcal{D}^2$-TimeAttn & 64.4\% & 87.5\% & 80.5\% & 75.5\% \\
\bottomrule
\end{tabular}
\end{table}

\begin{table}[tbp]
\centering
\caption{\textbf{Expected FLOPs (MFLOPs) per sample.} Computed analytically using the architectures in Appendix~\ref{app:probe_architectures}, with denoising steps $S$, hidden dimension $D$, and (for our cascade) average escalation rate $p_{\text{esc}}$ and average hesitation window $S_{\text{win}}$ measured on \textit{WildGuardMix}.}
\label{tab:flops}
\resizebox{\textwidth}{!}{%
\begin{tabular}{lcccc}
\toprule
& \textbf{LLaDA-8B-Base} & \textbf{LLaDA-8B-Instruct} & \textbf{LLaDA-1.5} & \textbf{LLaDA-2.0-mini} \\
\textbf{Method} & FLOPs & FLOPs & FLOPs & FLOPs \\
\midrule
\textit{Single-step methods} \\
\quad LP (Last Step) & 0.008 & 0.008 & 0.008 & 0.004 \\
\quad MLP (Last Step) & 2.10 & 2.10 & 2.10 & 1.05 \\
\midrule
\textit{Full-trajectory methods} \\
\quad LP (MV) & 0.26 & 0.26 & 0.26 & 0.52 \\
\quad LP (Mean) & 0.14 & 0.14 & 0.14 & 0.27 \\
\quad MLP (MV) & 67.1 & 67.1 & 67.1 & 134 \\
\quad MLP (Mean) & 2.23 & 2.23 & 2.23 & 1.31 \\
\quad TimeAttn & 35.7 & 35.7 & 35.7 & 68.2 \\
\quad LSTM & 163 & 163 & 163 & 386 \\
\midrule
\rowcolor{aliceblue}
$\boldsymbol{\mathcal{D}^2}$\textbf{-MLP (Ours)} & 0.92 & 0.68 & 0.97 & 0.87 \\
\rowcolor{aliceblue}
$\boldsymbol{\mathcal{D}^2}$\textbf{-TimeAttn (Ours)} & 8.16 & 5.10 & 8.48 & 19.4 \\
\bottomrule
\end{tabular}
}

\end{table}

\begin{table*}[tbp]
\centering
\normalsize
\renewcommand{\arraystretch}{1.4}
\setlength{\tabcolsep}{8pt}
\caption{\textbf{Post-generation inference time (ms).} Results are measured on WildGuardMix (1,725 prompts) after generation on different LLaDA models.}
\label{tab:inference_time_all}
\resizebox{\textwidth}{!}{%
\begin{tabular}{l c cc cc cc c cc}
\toprule
& & \multicolumn{2}{c}{\textbf{LLaDA-8B-Base}} & \multicolumn{2}{c}{\textbf{LLaDA-8B-Instruct}} & \multicolumn{2}{c}{\textbf{LLaDA-1.5}} & & \multicolumn{2}{c}{\textbf{LLaDA-2.0-mini}} \\
\cmidrule(lr){3-4} \cmidrule(lr){5-6} \cmidrule(lr){7-8} \cmidrule(lr){10-11}
\textbf{Method} & & Time &  & Time &  & Time &  &  & Time &  \\
\midrule

\textit{Single-step methods} \\[2pt]
\quad LP (Last Step)   & & 0.73 &  & 0.31 &  & 0.42 &  &  & 0.37 &  \\
\quad MLP (Last Step)  & & 0.79 &  & 0.52 &  & 0.66 &  &  & 0.34 &  \\

\midrule

\textit{Full-trajectory methods} \\[2pt]
\quad LP (MV)          & & 0.61 &  & 0.42 &  & 0.36 &  &  & 0.44 &  \\
\quad LP (Mean)        & & 2.46 &  & 1.78 &  & 1.10 &  &  & 3.33 &  \\
\quad MLP (MV)         & & 0.63 &  & 0.68 &  & 0.71 &  &  & 0.43 &  \\
\quad MLP (Mean)       & & 3.04 &  & 1.94 &  & 1.59 &  &  & 3.65 &  \\
\quad TimeAttn         & & 30.96 &  & 33.34 &  & 31.61 &  &  & 26.31 &  \\
\quad LSTM             & & 346.44 &  & 317.42 &  & 307.88 &  &  & 348.56 &  \\

\midrule

\rowcolor{aliceblue}
\quad $\boldsymbol{\mathcal{D}^2}$\textbf{-MLP (Ours)}        & & 0.57 &  & 0.56 &  & 0.66 &  &  & 0.55 &  \\
\rowcolor{aliceblue}
\quad $\boldsymbol{\mathcal{D}^2}$\textbf{-TimeAttn (Ours)}   & & 7.36 &  & 7.69 &  & 7.63 &  &  & 5.11 &  \\

\bottomrule
\end{tabular}%
}
\end{table*}

\begin{table}[tbp]
\centering
\normalsize
\renewcommand{\arraystretch}{1.4}
\setlength{\tabcolsep}{8pt}
\caption{\textbf{Intra-dataset performance on LLaDA-8B-Base.} Monitors are trained and tested on \textit{WildGuardMix}. Results are mean$\pm$std over 3 seeds. Best results are in \textbf{bold}, and second-best are \underline{underlined}. $\downarrow$ indicates lower is better.}
\label{tab:wgm_llada_8b_base}
\resizebox{\textwidth}{!}{%
\begin{tabular}{lccccccc}
\toprule
\textbf{Method} & \textbf{E[P]} & \textbf{Acc} & \textbf{F1} & \textbf{F2} & \textbf{Precision} & \textbf{Recall} & \textbf{FRR} $\downarrow$ \\
\midrule
\textit{Single-step methods} \\
\quad LP (Last Step)   & $4\!\times\!10^{-3}$M & 84.2$\pm$0.2 & 83.7$\pm$0.2 & 79.3$\pm$0.6 & 84.5$\pm$0.7 & 78.1$\pm$0.8 & 11.1$\pm$0.6 \\
\quad MLP (Last Step)  & 1.05M                 & 85.6$\pm$0.3 & 85.2$\pm$0.4 & 80.2$\pm$0.5 & 87.2$\pm$0.4 & 78.6$\pm$0.5 & 9.0$\pm$0.2 \\
\midrule
\textit{Full-trajectory methods} \\
\quad LP (MV)          & $4\!\times\!10^{-3}$M & 85.2$\pm$0.7 & 84.8$\pm$0.7 & 79.8$\pm$0.9 & 86.6$\pm$1.0 & 78.3$\pm$0.9 & 9.4$\pm$0.7 \\
\quad LP (Mean)        & $4\!\times\!10^{-3}$M & 85.7$\pm$1.0 & 85.3$\pm$1.0 & 80.7$\pm$1.2 & 87.0$\pm$1.2 & 79.3$\pm$1.2 & 9.2$\pm$0.9 \\
\quad MLP (MV)         & 1.05M                 & 86.8$\pm$0.4 & 86.4$\pm$0.4 & 81.4$\pm$0.9 & 89.0$\pm$0.2 & 79.8$\pm$1.1 & 7.7$\pm$0.3 \\
\quad MLP (Mean)       & 1.05M                 & 87.5$\pm$0.3 & 87.1$\pm$0.4 & 82.1$\pm$1.0 & 89.9$\pm$0.4 & 80.4$\pm$1.3 & 7.0$\pm$0.4 \\
\quad TimeAttn         & 1.59M                 & 87.6$\pm$0.5 & 87.2$\pm$0.6 & 82.8$\pm$3.0 & 89.4$\pm$1.6 & 81.3$\pm$3.3 & 7.5$\pm$1.0 \\
\quad LSTM             & 2.57M                 & 87.0$\pm$0.4 & 86.5$\pm$0.4 & 82.6$\pm$0.6 & 88.0$\pm$0.2 & 81.3$\pm$0.7 & 8.6$\pm$0.2 \\
\midrule
\rowcolor{aliceblue} \quad $\boldsymbol{\mathcal{D}^2}$\textbf{-MLP (Ours)}      & 0.34$\pm$0.06M & \underline{88.5$\pm$0.2} & \underline{88.2$\pm$0.3} & \underline{83.6$\pm$0.5} & \underline{90.9$\pm$0.0} & \underline{81.9$\pm$0.6} & \underline{6.4$\pm$0.1} \\
\rowcolor{aliceblue} \quad $\boldsymbol{\mathcal{D}^2}$\textbf{-TimeAttn (Ours)} & 0.55$\pm$0.03M & \textbf{89.4$\pm$0.3} & \textbf{89.1$\pm$0.4} & \textbf{85.1$\pm$0.4} & \textbf{91.3$\pm$0.5} & \textbf{83.7$\pm$0.4} & \textbf{6.2$\pm$0.4} \\
\bottomrule
\end{tabular}%
}
\end{table}
\begin{table}[tbp]
\centering
\normalsize
\renewcommand{\arraystretch}{1.4}
\setlength{\tabcolsep}{8pt}
\caption{\textbf{Intra-dataset performance on LLaDA-8B-Instruct.} Monitors are trained and tested on \textit{WildGuardMix}. Results are mean$\pm$std over 3 seeds. Best results are in \textbf{bold}, and second-best are \underline{underlined}. $\downarrow$ indicates lower is better.}
\label{tab:wgm_llada_8b_instruct}
\resizebox{\textwidth}{!}{%
\begin{tabular}{lccccccc}
\toprule
\textbf{Method} & \textbf{E[P]} & \textbf{Acc} & \textbf{F1} & \textbf{F2} & \textbf{Precision} & \textbf{Recall} & \textbf{FRR} $\downarrow$ \\
\midrule
\textit{Single-step methods} \\
\quad LP (Last Step)   & $4\!\times\!10^{-3}$M & 87.0$\pm$0.3 & 86.6$\pm$0.3 & 83.1$\pm$0.2 & 87.3$\pm$1.0 & 82.1$\pm$0.5 & 9.3$\pm$0.9 \\
\quad MLP (Last Step)  & 1.05M                 & 87.2$\pm$0.5 & 86.9$\pm$0.5 & 83.5$\pm$0.5 & 87.4$\pm$0.7 & 82.5$\pm$0.5 & 9.2$\pm$0.6 \\
\midrule
\textit{Full-trajectory methods} \\
\quad LP (MV)          & $4\!\times\!10^{-3}$M & 87.8$\pm$0.5 & 87.6$\pm$0.5 & 84.8$\pm$0.7 & 87.5$\pm$0.7 & 84.2$\pm$0.8 & 9.3$\pm$0.5 \\
\quad LP (Mean)        & $4\!\times\!10^{-3}$M & 87.9$\pm$0.4 & 87.6$\pm$0.4 & 84.4$\pm$0.7 & 88.2$\pm$0.8 & 83.5$\pm$1.0 & 8.7$\pm$0.7 \\
\quad MLP (MV)         & 1.05M                 & 88.0$\pm$0.4 & 87.7$\pm$0.4 & 84.1$\pm$0.8 & 88.8$\pm$0.1 & 83.0$\pm$1.0 & 8.1$\pm$0.2 \\
\quad MLP (Mean)       & 1.05M                 & 88.1$\pm$0.4 & 87.8$\pm$0.4 & 84.2$\pm$0.7 & 89.0$\pm$0.1 & 83.1$\pm$0.9 & 8.0$\pm$0.1 \\
\quad TimeAttn         & 1.59M                 & 88.1$\pm$0.2 & 87.8$\pm$0.2 & 84.3$\pm$0.7 & 88.9$\pm$1.3 & 83.2$\pm$1.1 & 8.1$\pm$1.1 \\
\quad LSTM             & 2.57M                 & 88.2$\pm$0.4 & 87.9$\pm$0.4 & 84.6$\pm$1.1 & 88.7$\pm$1.5 & 83.6$\pm$1.5 & 8.3$\pm$1.3 \\
\midrule
\rowcolor{aliceblue} \quad $\boldsymbol{\mathcal{D}^2}$\textbf{-MLP (Ours)}      & 0.26$\pm$0.03M & \textbf{89.8$\pm$0.1} & \textbf{89.5$\pm$0.1} & \textbf{86.1$\pm$0.3} & \underline{91.2$\pm$0.3} & \textbf{84.9$\pm$0.5} & \underline{6.4$\pm$0.3} \\
\rowcolor{aliceblue} \quad $\boldsymbol{\mathcal{D}^2}$\textbf{-TimeAttn (Ours)} & 0.22$\pm$0.04M & \underline{89.6$\pm$0.1} & \underline{89.3$\pm$0.1} & \underline{85.6$\pm$0.6} & \textbf{91.2$\pm$0.6} & \underline{84.3$\pm$0.8} & \textbf{6.3$\pm$0.5} \\
\bottomrule
\end{tabular}%
}
\end{table}
\clearpage
\begin{table}[tbp]
\centering
\normalsize
\renewcommand{\arraystretch}{1.4}
\setlength{\tabcolsep}{8pt}
\caption{\textbf{Intra-dataset performance on LLaDA-1.5.} Monitors are trained and tested on \textit{WildGuardMix}. Results are mean$\pm$std over 3 seeds. Best results are in \textbf{bold}, and second-best are \underline{underlined}. $\downarrow$ indicates lower is better.}
\label{tab:wgm_llada_15}
\resizebox{\textwidth}{!}{%
\begin{tabular}{lccccccc}
\toprule
\textbf{Method} & \textbf{E[P]} & \textbf{Acc} & \textbf{F1} & \textbf{F2} & \textbf{Precision} & \textbf{Recall} & \textbf{FRR} $\downarrow$ \\
\midrule
\textit{Single-step methods} \\
\quad LP (Last Step) & $4\!\times\!10^{-3}$M & 88.2$\pm$0.1 & 87.9$\pm$0.1 & 83.9$\pm$0.1 & 89.6$\pm$0.3 & 82.6$\pm$0.2 & 7.5$\pm$0.3 \\
\quad MLP (Last Step)  & 1.05M                 & 87.3$\pm$0.0 & 87.0$\pm$0.0 & 83.9$\pm$0.1 & 87.3$\pm$0.2 & 83.1$\pm$0.2 & 9.4$\pm$0.2 \\
\midrule
\textit{Full-trajectory methods} \\
\quad LP (MV)          & $4\!\times\!10^{-3}$M & 87.4$\pm$0.1 & 87.2$\pm$0.1 & 84.7$\pm$0.2 & 86.7$\pm$0.1 & 84.2$\pm$0.3 & 10.1$\pm$0.1 \\
\quad LP (Mean)        & $4\!\times\!10^{-3}$M & 87.7$\pm$0.1 & 87.4$\pm$0.1 & 84.9$\pm$0.1 & 87.0$\pm$0.2 & 84.4$\pm$0.2 & 9.8$\pm$0.2 \\
\quad MLP (MV)         & 1.05M                 & 87.8$\pm$0.1 & 87.5$\pm$0.0 & 84.1$\pm$0.5 & 88.3$\pm$0.7 & 83.2$\pm$0.8 & 8.5$\pm$0.7 \\
\quad MLP (Mean)       & 1.05M                 & 88.1$\pm$0.0 & 87.8$\pm$0.0 & 84.2$\pm$0.5 & 88.9$\pm$0.7 & 83.1$\pm$0.8 & 8.0$\pm$0.6 \\
\quad TimeAttn         & 1.59M                 & 88.2$\pm$0.4 & 88.0$\pm$0.5 & 84.1$\pm$1.0 & 90.0$\pm$0.2 & 82.8$\pm$1.3 & 7.1$\pm$0.2 \\
\quad LSTM             & 2.57M                 & 88.1$\pm$0.4 & 87.9$\pm$0.4 & 84.5$\pm$0.7 & 88.6$\pm$0.3 & 83.6$\pm$0.9 & 8.3$\pm$0.2 \\
\midrule
\rowcolor{aliceblue} \quad $\boldsymbol{\mathcal{D}^2}$\textbf{-MLP (Ours)}      & 0.27$\pm$0.08M & \underline{89.7$\pm$0.1} & \underline{89.4$\pm$0.1} & \textbf{86.1$\pm$0.3} & \underline{90.7$\pm$0.2} & \textbf{85.1$\pm$0.4} & \underline{6.8$\pm$0.2} \\
\rowcolor{aliceblue} \quad $\boldsymbol{\mathcal{D}^2}$\textbf{-TimeAttn (Ours)} & 0.37$\pm$0.07M & \textbf{89.7$\pm$0.2} & \textbf{89.4$\pm$0.2} & \underline{86.0$\pm$0.2} & \textbf{90.9$\pm$0.4} & \underline{84.9$\pm$0.1} & \textbf{6.6$\pm$0.3} \\
\bottomrule
\end{tabular}%
}
\end{table}
\begin{table}[tbp]
\centering
\normalsize
\renewcommand{\arraystretch}{1.4}
\setlength{\tabcolsep}{8pt}
\caption{\textbf{Intra-dataset performance on LLaDA-2.0-mini.} Monitors are trained and tested on \textit{WildGuardMix}. Results are mean$\pm$std over 3 seeds. Best results are in \textbf{bold}, and second-best are \underline{underlined}. $\downarrow$ indicates lower is better.}
\label{tab:wgm_llada_20_mini}
\resizebox{\textwidth}{!}{%
\begin{tabular}{lccccccc}
\toprule
\textbf{Method} & \textbf{E[P]} & \textbf{Acc} & \textbf{F1} & \textbf{F2} & \textbf{Precision} & \textbf{Recall} & \textbf{FRR} $\downarrow$ \\
\midrule
\textit{Single-step methods} \\
\quad LP (Last Step) & $2\!\times\!10^{-3}$M & 77.4$\pm$0.2 & 76.1$\pm$0.2 & 65.2$\pm$0.4 & 81.8$\pm$0.5 & 62.1$\pm$0.4 & 10.7$\pm$0.3 \\
\quad MLP (Last Step)  & 0.52M                 & 78.5$\pm$0.1 & 77.6$\pm$0.1 & 68.9$\pm$0.1 & 81.1$\pm$0.3 & 66.4$\pm$0.2 & 12.0$\pm$0.3 \\
\midrule
\textit{Full-trajectory methods} \\
\quad LP (MV)          & $2\!\times\!10^{-3}$M & 80.1$\pm$0.3 & 79.1$\pm$0.2 & 69.7$\pm$0.5 & 84.3$\pm$1.3 & 66.8$\pm$0.8 & 9.6$\pm$1.1 \\
\quad LP (Mean)        & $2\!\times\!10^{-3}$M & 80.3$\pm$0.2 & 79.4$\pm$0.2 & 70.4$\pm$1.5 & 84.4$\pm$1.9 & 67.6$\pm$2.0 & 9.8$\pm$1.7 \\
\quad MLP (MV)         & 0.52M                 & 80.9$\pm$0.1 & 80.0$\pm$0.0 & 71.3$\pm$1.3 & 84.9$\pm$1.8 & 68.6$\pm$1.8 & 9.5$\pm$1.5 \\
\quad MLP (Mean)       & 0.52M                 & 81.2$\pm$0.1 & 80.4$\pm$0.0 & 71.9$\pm$1.0 & 85.1$\pm$1.4 & 69.2$\pm$1.3 & 9.4$\pm$1.2 \\
\quad TimeAttn         & 0.80M                 & 81.7$\pm$0.4 & 80.7$\pm$0.2 & 70.6$\pm$1.2 & \underline{88.1$\pm$2.6} & 67.2$\pm$1.7 & \underline{7.3$\pm$2.0} \\
\quad LSTM             & 1.51M                 & 81.7$\pm$0.6 & 80.8$\pm$0.5 & 71.4$\pm$0.6 & 87.1$\pm$2.2 & 68.3$\pm$1.0 & 7.9$\pm$1.7 \\
\midrule
\rowcolor{aliceblue} \quad $\boldsymbol{\mathcal{D}^2}$\textbf{-MLP (Ours)}      & 0.06$\pm$0.01M & \underline{84.0$\pm$0.1} & \underline{83.4$\pm$0.1} & \textbf{76.4$\pm$0.3} & 87.4$\pm$0.4 & \textbf{74.0$\pm$0.4} & 8.3$\pm$0.4 \\
\rowcolor{aliceblue} \quad $\boldsymbol{\mathcal{D}^2}$\textbf{-TimeAttn (Ours)} & 0.12$\pm$0.04M & \textbf{84.2$\pm$0.3} & \textbf{83.5$\pm$0.3} & \underline{75.7$\pm$0.7} & \textbf{88.7$\pm$0.5} & \underline{73.1$\pm$0.8} & \textbf{7.2$\pm$0.4} \\
\bottomrule
\end{tabular}%
}
\end{table}

\clearpage

\clearpage




\newpage

\end{document}